\documentclass[lettersize,journal]{IEEEtran}
\usepackage{caption}
\usepackage{array}
\usepackage[caption=false,font=normalsize,labelfont=sf,textfont=sf]{subfig}
\usepackage{textcomp}
\usepackage{stfloats}
\usepackage{url}
\usepackage{hyperref}
\usepackage{breqn}
\usepackage{verbatim}
\usepackage{graphicx}
\usepackage{cite}
\usepackage{mdframed}
\usepackage{amsmath,amssymb,amsfonts}
\usepackage{amsthm}
\theoremstyle{remark}

\usepackage{graphicx}
\usepackage{textcomp}

\usepackage{flushend}
\usepackage{balance}
\usepackage{tcolorbox}
\tcbuselibrary{skins, breakable}

\newtcolorbox{notebox}{
  enhanced,
  breakable,
  colback=white,
  colframe=black!50,
  arc=2pt,
  boxrule=0.4pt,
  left=6pt, right=6pt, top=4pt, bottom=4pt,
  fontupper=\small\itshape
}

\usepackage{tikz}
\usetikzlibrary{positioning, fit, backgrounds, arrows.meta}
\usepackage[table,xcdraw]{xcolor}
\usepackage[linesnumbered,ruled,lined]{algorithm2e}

\newenvironment{tightalgorithm}
  {%
   \setlength{\textfloatsep}{5pt}%
   \setlength{\intextsep}{5pt}%
   \begin{algorithm}
  }
  {%
   \end{algorithm}
  }
\newcounter{parnum}[subsubsection] % reset every subsection
\renewcommand{\theparnum}{\alph{parnum}} % use a, b, c...
\newcommand{\parletter}[1]{%
  \refstepcounter{parnum}%
  \noindent\textbf{\theparnum) #1}\par
}

\makeatletter
\let\@oldmaketitle\@maketitle
\renewcommand{\@maketitle}{\@oldmaketitle
\centering
 \includegraphics[width=\linewidth]{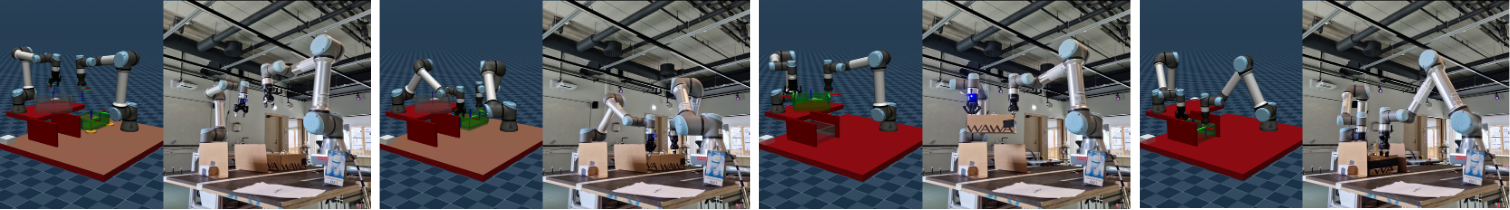}
\setcounter{figure}{0}
\captionof{figure}{\footnotesize{Simulation and real-world execution of PRISM. Two UR5e arms cooperatively pick up a tray and navigate through a cluttered workspace to place it between two obstacles. Left-to-right sequences demonstrate the system dynamically adapting its trajectory to maintain a level grasp during the move phase}}
\label{fig:moving_tray_snapshots}
\vspace{-0.8cm}
}

\makeatother

\allowdisplaybreaks
\begin{document}

\title{\LARGE \bf
PRISM: Projection-Integrated Sampling-Based MPC with Bayesian Cost Tuning for Bimanual Manipulation
}
\author{Alinjar Dan$^{1*}$, Iryna Hurova$^{1*}$, Karl Kruusamäe$^{1}$, Arun Kumar Singh$^{1}$% <-this % stops a space
\thanks{$^{*}$These authors contributed equally to this work.}
\thanks{$^{1}$ University of Tartu, Narva mnt 18, 51009, Tartu, Estonia\\
\vspace{1mm}
Alinjar Dan: Post-doc, \url{alinjardannitdgp2014@gmail.com}\\
Iryna Hurova: Post-graduate student, \url{iryna.gurova@gmail.com}\\
Karl Kruusamäe: Associate Professor, \url{karl.kruusamae@ut.ee}\\
Arun Kumar Singh: Associate Professor, \url{aks1812@gmail.com}}
}

\maketitle

\begin{abstract}
Bimanual manipulation in cluttered, contact-rich environments remains challenging because it requires coordinated motion generation, interaction-aware planning, and reliable execution under tight kinematic constraints. We present PRISM, a projection-integrated sampling-based Model Predictive Control (MPC) framework that uses a GPU-accelerated physics simulator as an online world model for complex dual-arm manipulation.

The main algorithmic contribution is a QP-guided control sampling strategy that decouples trajectory exploration from kinematic feasibility. At each MPC step, sampled joint-velocity trajectories are projected onto the set of motions satisfying joint position, velocity, acceleration, and jerk bounds, together with an initial-velocity boundary condition, before rollout evaluation. This enables broad yet feasible exploration of coordinated bimanual behaviors. To support efficient online execution, we derive a custom ADMM/Bregman-splitting QP solver that exploits joint-wise separability and reusable matrix factorizations. We further use Bayesian optimization to tune task-cost weights offline, reducing manual parameter selection.

We evaluate PRISM on challenging variants of PerAct$^{2}$ tasks, including obstacle-constrained ball transport, tray transport, cube handover, and box lifting. Experiments show improved robustness and task success relative to representative sampling-based baselines, while maintaining real-time or near-real-time execution. We also demonstrate successful sim-to-real transfer on dual UR5e manipulators, highlighting the practical potential of physics-based online planning for contact-rich bimanual manipulation.
Project details, including code and supplementary videos, are available at \href{https://sites.google.com/view/prismbimanual}{\texttt{https://sites.google.com/view/prismbimanual}}.
\end{abstract}

\def\abstractname{Note to Practitioners}

\begin{abstract}
This paper addresses the problem of programming dual-arm robots to reliably manipulate and transport objects in cluttered workspaces, with applications in assembly and logistics. Rather than relying on task-specific policies learned from large demonstration datasets, our framework uses a GPU-accelerated physics simulator as an online world model to evaluate many candidate bimanual motions in parallel at each control step and select a feasible action sequence for the current scene. A key practical feature is that smooth, executable motions are enforced directly during the search process: sampled trajectories are projected onto a set satisfying joint position, velocity, acceleration, and jerk limits before rollout evaluation. This enables broad exploration while avoiding highly oscillatory or dynamically inconsistent commands, which is beneficial for real robot execution. The task-cost weights are tuned offline using Bayesian optimization, reducing manual parameter adjustment without increasing online control cost. The approach requires a reasonably accurate physics model of the robot and workspace, and the task cost structure must still be designed by the user; within these assumptions, it provides a practical alternative to demonstration-heavy policy learning for contact-rich bimanual manipulation and related constraint-aware robotic planning problems.
\end{abstract}

\begin{IEEEkeywords}
Bimanual, Sampling based MPC, Bayesian Optimization, Quadratic programming, MuJoCo.
\end{IEEEkeywords}

\section{Introduction}
Dual-arm manipulation represents a significant frontier in robotics, holding the promise of emulating human-level dexterity for complex tasks such as intricate assembly, collaborative object handling, and dynamic interaction with unstructured environments. The ability to coordinate two robotic arms unlocks capabilities that are difficult or impossible for a single manipulator, including stabilizing large objects, performing simultaneous actions, and applying opposing forces. This potential has fueled a surge of research interest, with the goal of developing autonomous systems that can operate robustly and intelligently in the physical world.
 
\begin{figure*}
   \centering
   \includegraphics[width=0.8\linewidth]{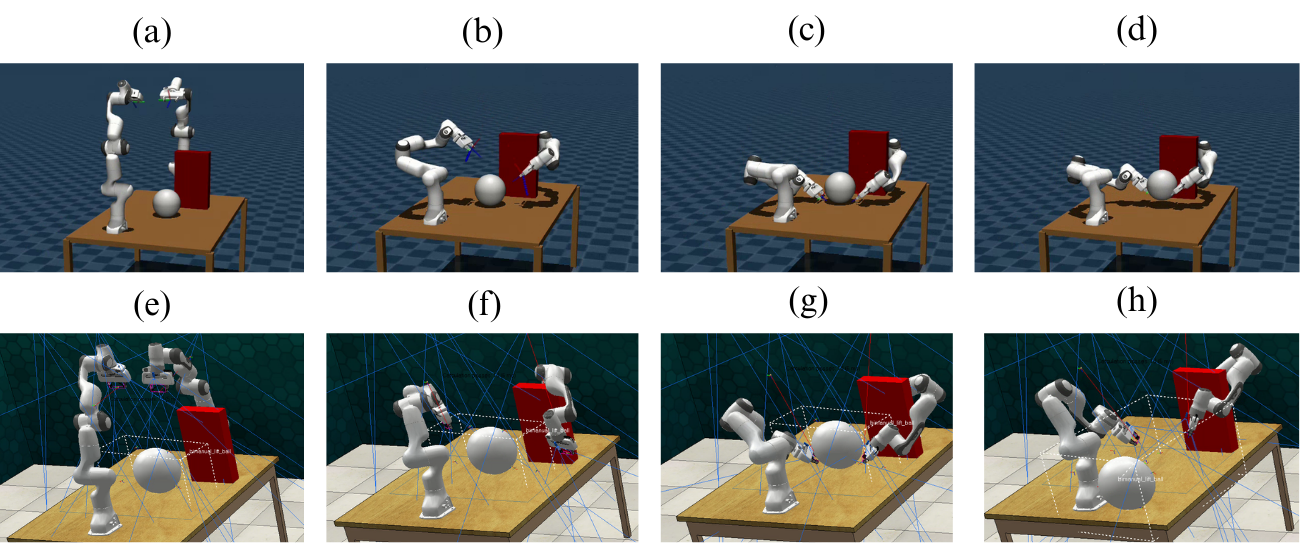}
   \caption{{\footnotesize{Illustrative comparison of our 
approach (top) with~\cite{gkanatsios20253d} (bottom) on a 
ball-lifting task with an added obstacle. In the obstacle-free 
setting, both approaches succeed. When an obstacle is introduced---a 
configuration outside the training distribution 
of~\cite{gkanatsios20253d}---the learned policy struggles to 
adapt its lifting strategy, while our online planner discovers a 
feasible solution by reasoning over the current scene.}}}
\label{fig:teaser}
\end{figure*}

In recent years, the dominant paradigm for tackling bimanual 
manipulation has been end-to-end 
learning~\cite{lu2024anybimanual, gkanatsios20253d, grotz2024peract2, 
black2024pi_0}. Approaches based on imitation learning and 
reinforcement learning have demonstrated remarkable success in 
learning complex, high-dimensional policies directly from data. 
Their popularity stems from two key strengths. First, they can 
directly map high-dimensional perceptual input to motor commands. 
Second, they bypass the need to explicitly model the interactions 
between the arms and between the arms and the objects they 
manipulate. For example, consider the task of lifting a ball with 
both arms: by learning directly from demonstration data, imitation 
learning avoids the need to explicitly model the contact forces 
between the ball and the end-effectors. However, the success of 
these approaches is predicated on the availability of sufficiently 
diverse training data and remains tightly coupled to the 
distribution of demonstrated scenarios. When faced with 
environmental configurations not well-represented in the training 
set---such as novel obstacle placements or modified task 
constraints---learned policies may require additional data 
collection or fine-tuning to recover reliable performance 
(see Fig.~\ref{fig:teaser} for an illustrative example). This 
sensitivity to distribution shift motivates the exploration of 
alternative planning approaches that can adapt online to 
previously unseen configurations.

% === Replace "Our Approach and Its Novelty" paragraph ===

In this work, we investigate a complementary paradigm: real-time, model-based planning with a GPU-accelerated physics simulator serving as an online world model. Rather than learning a monolithic task policy, our approach uses massively parallel physics rollouts to evaluate large batches of candidate bimanual motion sequences at every planning step. This allows the planner to directly reason over the current geometric and physical state of the scene, including previously unseen obstacle layouts, object poses, and contact interactions. Such an approach is particularly attractive for contact-rich bimanual manipulation, where collecting sufficiently diverse expert demonstrations is itself challenging and where task success depends on adapting motions online to the evolving interaction dynamics.

\subsubsection*{Our Algorithmic Contribution}
We present PRISM, a projection-integrated sampling-based Model Predictive Control (MPC) framework, augmented with Bayesian optimization, for complex bimanual manipulation tasks. PRISM combines GPU-parallel physics-based rollout evaluation with a Quadratic Program(QP)-guided control sampling strategy that decouples trajectory exploration from kinematic feasibility. At each MPC step, the planner samples joint-velocity trajectories from an adaptive Gaussian distribution, projects every sample onto the set of trajectories satisfying joint position, velocity, acceleration, and jerk bounds together with MPC boundary-continuity constraints, and evaluates the resulting feasible rollouts in parallel through the GPU-accelerated MuJoCo world model. Because feasibility is enforced \emph{before} rollout, rather than introduced through post-processing or soft penalties, the optimizer can employ aggressive exploration covariances while ensuring that all candidate motions remain smooth and executable. The projected samples are then importance-weighted and used to update both the sampling mean and covariance, yielding a receding-horizon optimizer that progressively concentrates on high-quality coordinated bimanual behaviors.

\subsubsection*{Benefits of our Approach}
We apply PRISM to challenging variants of tasks from the PerAct$^{2}$ benchmark~\cite{grotz2024peract2}. For example, we extend a standard ball-lifting task into a longer-horizon transport problem in which the object must be moved through an obstacle-constrained workspace. These settings are especially demanding for learned policies when the required avoidance strategies are not represented in the training distribution. Across ball transport, tray transport, cube handover, and box lifting, we show that PRISM outperforms representative sampling-based alternatives such as~\cite{kicki2025low,williams2018information,baseline_cem} in robustness, task success, and execution efficiency in the considered bimanual settings. We further quantify the gains provided by Bayesian optimization of the cost weights and demonstrate successful sim-to-real transfer on dual UR5e manipulators, including cluttered manipulation scenarios. Finally, we provide a preliminary qualitative comparison with~\cite{gkanatsios20253d} to illustrate the complementary strength of online planning under geometric distribution shift.

\vspace{-0.4cm}
\section{RELATED WORK}

We organize the relevant literature along four axes: sampling-based planning for manipulation, smoothness in sampling-based MPC, automatic cost-function tuning, and end-to-end learning for bimanual manipulation.

\subsection{Sampling-Based Planning for Manipulation}

\noindent Gradient-based trajectory optimization has been applied to bimanual tasks such as synchronous pick-and-place~\cite{martinez2025trajectory} and cooperative object transportation~\cite{karim2025vil, hu2022adaptive}. These methods can be highly effective when the task geometry and contact structure are well characterized. However, extending them to cluttered settings or manipulation problems involving the making and breaking of contacts often requires substantial task-specific modeling. In contrast, sampling-based optimization~\cite{howell2022predictive, pezzato2025sampling} requires only the ability to evaluate forward rollouts of a world model and arbitrary task costs. This flexibility has made sampling-based MPC increasingly attractive for contact-rich manipulation, particularly as GPU-accelerated physics simulators such as MuJoCo~\cite{howell2022predictive} and IsaacGym~\cite{pezzato2025sampling} enable massively parallel trajectory evaluation.

Among sampling-based approaches, Howell et al.~\cite{howell2022predictive} demonstrated preliminary bimanual manipulation results using predictive sampling with MuJoCo, while Pezzato et al.~\cite{pezzato2025sampling} scaled sampling-based MPC through parallelized physics simulation in IsaacGym. A central difficulty in extending such methods to complex bimanual manipulation is the presence of coordination constraints and closed-chain-like interaction structures. Classical motion-planning theory emphasizes that exact closed-chain constraints define lower-dimensional manifolds that are difficult to explore using naive random sampling~\cite{acar2021approximating}. Traditional control approaches address this issue through carefully designed boundary-layer or noninteracting controllers~\cite{bonilla2017noninteracting}, while more recent learning-based methods use generative models such as CVAEs, CGANs~\cite{acar2021approximating}, or energy-based samplers~\cite{tong2025adaptive} to draw constraint-aware samples offline. However, such approaches typically separate task-constraint learning from online environmental reasoning, and their applicability to contact-rich manipulation with changing obstacle configurations remains limited.

\begin{notebox}
    Our Improvement: Our approach  discovers coordinated dual-arm behavior by exploring directly in the combined joint space, while coordination requirements are expressed through task costs evaluated in the physics world model. Building on the predictive-sampling paradigm of~\cite{howell2022predictive}, we introduce a substantially more structured optimizer that enforces jerk-bounded smoothness on all sampled trajectories and permits higher exploration noise. Unlike approaches that rely on offline learned constraint samplers, our method handles contact-rich manipulation online, runs at $10$--$16$\,Hz in the considered settings, and is validated in both simulation and real-world experiments.
\end{notebox}

\subsection{Smoothness in Sampling-Based MPC}

\noindent A well-known limitation of sampling-based MPC algorithms such as MPPI~\cite{williams2017model} and CEM~\cite{baseline_cem} is that sampled control sequences can be highly non-smooth. This may lead to oscillatory actuation, degraded warm-starting across MPC iterations, and poor task performance in systems with physical bandwidth or actuator limits. Several strategies have been proposed to address this issue.

\paragraph{Post-hoc filtering}
A common approach is to smooth the selected control sequence after the optimizer update, for example using a Savitzky--Golay filter (SGF)~\cite{williams2018information}. While simple to implement, post-hoc filtering does not alter the trajectories evaluated during optimization and provides no guarantee that the resulting controls satisfy actuator or derivative bounds.

\paragraph{Control-space lifting}
Kim et al.~\cite{kim2022smooth} proposed sampling in the space of control derivatives while treating controls as additional state variables. This shifts smoothness handling into the optimization process through augmented dynamics and penalty terms. However, the method introduces additional design choices and cost weights, and these difficulties become more pronounced when higher-order derivative limits are required.

\paragraph{Frequency-domain shaping}
Vlahov et al.~\cite{vlahov2024low} introduced colored-noise sampling for MPPI, biasing perturbations toward low-frequency components. More recently, Kicki~\cite{kicki2025low} proposed LPF-MPPI, which applies a Butterworth low-pass filter to sampled perturbations before rollout. These methods provide direct control over the spectral content of exploration noise, but they do not enforce strict bounds on control derivatives. A low-frequency sample can still violate acceleration or jerk limits if its amplitude is sufficiently large. In addition, such methods do not directly impose boundary conditions needed for continuity across MPC replanning steps.

\paragraph{Polynomial and spline parameterization}
STORM~\cite{bhardwaj2022storm} and SCP-MPPI~\cite{miura2024spline} represent control sequences through polynomial bases or spline interpolation of a reduced set of parameters. These formulations induce smoothness by restricting representational bandwidth. However, the level of smoothness is controlled only indirectly through the parameterization, and strict derivative bounds are generally not enforced.

\paragraph{Projection-based methods}
Andrejev et al.~\cite{andrejev2025pi} introduced $\pi$-MPPI, which incorporates a QP-based projection step into the MPPI pipeline to minimally correct sampled controls so that they satisfy bounds on control magnitudes and derivatives. This projection decouples the exploration distribution from feasibility: the sampling covariance may be chosen aggressively, while the projection maps each sample to the closest admissible trajectory. Rastgar et al.~\cite{rastgar2024priest} also explored projection-guided sampling for autonomous navigation, although without the derivative-bounded formulation used in $\pi$-MPPI.

\begin{notebox}
 Our Improvement: We adapt the projection-based principle of~\cite{andrejev2025pi} to the bimanual joint-velocity setting, extend the derivative constraints to include jerk bounds up to $r=3$, and derive a custom ADMM solver that exploits the joint-wise separability of the resulting QP for efficient GPU parallelization. As shown in Section~\ref{subsec:Role of QP}, this projection-augmented sampling strategy enables substantially larger effective exploration covariances than vanilla sampling and frequency-domain alternatives, leading to improved success rates across the considered bimanual tasks.
\end{notebox}

\subsection{Automatic Cost Function Tuning}

\noindent The performance of sampling-based planners is highly sensitive to the relative weighting of competing objectives, including collision avoidance, task progress, bimanual coordination, and trajectory regularization. Manual tuning of these weights is labor-intensive, difficult to reproduce, and often specific to a particular task instance. Bayesian optimization (BO) has emerged as a principled tool for automated hyperparameter and reward tuning in robotics. Prior works have applied BO to tune MPC parameters~\cite{marco2016automatic} and to calibrate controller gains under safety constraints~\cite{berkenkamp2023bayesian}. However, automatic tuning of task-specific cost weights for multi-phase, multi-objective sampling-based MPC in bimanual manipulation has received comparatively limited attention.

\begin{notebox}
    Our Improvement: We use Bayesian optimization to tune the cost weights of our sampling-based planner offline, evaluating each candidate weight configuration over full task episodes with randomized initial conditions. This procedure improves success rates across all four evaluated tasks, as shown in Section~\ref{subsec:Bayesian_Optimization}, while introducing no additional computational overhead during online MPC execution.
\end{notebox}

\subsection{End-to-End Learning for Bimanual Manipulation}

\noindent End-to-end learning methods, including imitation learning and reinforcement learning, have demonstrated impressive performance on bimanual manipulation tasks by learning complex, high-dimensional policies directly from data~\cite{gkanatsios20253d, lu2024anybimanual, grotz2024peract2, black2024pi_0}. Recent imitation-learning systems often rely on expressive policy parameterizations based on diffusion models~\cite{janner2022planning} or flow matching~\cite{lipman2022flow}, which can represent multimodal action distributions and capture diverse manipulation strategies. Despite these advances, several challenges remain. First, collecting sufficiently diverse expert demonstrations is particularly difficult in cluttered manipulation settings. Second, neural policies do not naturally satisfy hard physical or kinematic constraints~\cite{dontidc3}. Third, the strongest diffusion- and flow-based policies may incur substantial inference cost~\cite{gkanatsios20253d}. Reinforcement learning can, in principle, reduce dependence on expert demonstrations, but existing approaches for bimanual manipulation are often task-specific~\cite{cui2024task} or rely on simplified manipulator models~\cite{chen2023bi}.

\begin{notebox}
    Our Improvement: PRISM offers a complementary model-based alternative. Rather than learning a fixed task policy, it adapts online to the current scene geometry through physics-based rollout evaluation. It explicitly reasons over obstacle configurations and contact-rich interactions, operates at $10$--$16$\,Hz in the considered settings, and can generate synthetic expert trajectories for future policy learning. More broadly, PRISM may serve as a useful planning component within model-based reinforcement-learning frameworks such as TD-MPC~\cite{hansen2022temporal}, combining online physical reasoning with learned long-horizon value information.
\end{notebox}

\section{METHODS}

Consider a bimanual setting where
$\boldsymbol{\theta}_{1,k}$ and $\boldsymbol{\theta}_{2,k}$
represent the joint-position vectors of the individual arms at planning step $k$.
Let
\[
\boldsymbol{\theta}_k
=
\left(
\boldsymbol{\theta}_{1,k},
\boldsymbol{\theta}_{2,k}
\right)
\]
denote the combined joint-position vector of the bimanual system. Let
$\mathbf{x}_k$ represent the environment state at step $k$. For example,
$\mathbf{x}_k$ can contain the Euclidean position of specific points on the manipulator bodies, as well as the position and velocity of movable objects in the environment. With these notations in place, we define bimanual planning over a horizon $H$ through the following optimization problem:
\begin{subequations}
\begin{align}
    \arg\min_{\dot{\boldsymbol{\theta}}_{1:H}}
    \quad
    &\sum_{k=1}^{H}
    c\!\left(
    \boldsymbol{\theta}_k,
    \dot{\boldsymbol{\theta}}_k,
    \mathbf{x}_k
    \right)
    \label{cost}\\
    \text{s.t.}\quad
    &
    \left(
    \boldsymbol{\theta}_{k+1},
    \mathbf{x}_{k+1}
    \right)
    =
    \mathbf{f}
    \left(
    \boldsymbol{\theta}_k,
    \dot{\boldsymbol{\theta}}_k,
    \mathbf{x}_k
    \right),
    \label{state_evol}\\
    &
    \boldsymbol{\theta}^{(r)}_{\min}
    \leq
    \boldsymbol{\theta}^{(r)}_k
    \leq
    \boldsymbol{\theta}^{(r)}_{\max},
    \qquad
    r\in\{0,1,2,3\},
    \label{bounds}\\
    &
    \boldsymbol{\theta}_0
    =
    \boldsymbol{\theta}_{\mathrm{init}},
    \qquad
    \dot{\boldsymbol{\theta}}_0
    =
    \dot{\boldsymbol{\theta}}_{\mathrm{init}}.
    \label{boundary}
\end{align}
\end{subequations}
Here, $\dot{\boldsymbol{\theta}}_{1:H}$ denotes the horizon-length sequence of joint-velocity vectors, and
$\boldsymbol{\theta}^{(r)}_k$ denotes the $r$-th time derivative of the joint trajectory at step $k$, with
$\boldsymbol{\theta}^{(0)}_k=\boldsymbol{\theta}_k$,
$\boldsymbol{\theta}^{(1)}_k=\dot{\boldsymbol{\theta}}_k$,
$\boldsymbol{\theta}^{(2)}_k=\ddot{\boldsymbol{\theta}}_k$,
and
$\boldsymbol{\theta}^{(3)}_k=\dddot{\boldsymbol{\theta}}_k$.

The stage cost is expressed as a weighted sum of task-specific components:
\begin{align}
    c\!\left(
    \boldsymbol{\theta}_k,
    \dot{\boldsymbol{\theta}}_k,
    \mathbf{x}_k
    \right)
    =
    \sum_i
    w_i\,
    \overline{c}_i
    \left(
    \boldsymbol{\theta}_k,
    \dot{\boldsymbol{\theta}}_k,
    \mathbf{x}_k
    \right).
    \label{cost_components}
\end{align}

\noindent
The objective in \eqref{cost} consists of several task-dependent cost components that model manipulation goals, coordination requirements, and collision avoidance. These components are weighted by coefficients $w_i$, which are later tuned through Bayesian optimization. We discuss their task-specific algebraic forms in Section~\ref{sec:cost_design}. The coupled joint-space and environment evolution is dictated by the function $\mathbf{f}$, which serves as our world model and is realized through the MuJoCo physics engine in \eqref{state_evol}. To ensure smooth joint motions and aid sim-to-real transfer, we enforce bounds on derivatives of the joint trajectory up to jerk level, as shown in \eqref{bounds}. The boundary conditions in \eqref{boundary} initialize the planning problem at the current joint configuration and enforce continuity with the current joint velocity at the MPC replanning boundary. It is worth emphasizing that the optimization solves for the motions of both arms jointly. Thus, an appropriate coordination strategy is discovered online for each task instance.

\begin{tightalgorithm}[!t]
\caption{Sampling-Based Optimizer for Solving \eqref{cost}--\eqref{bounds}}
\label{algo_1}
\small
\SetAlgoLined
\textbf{Input:} MuJoCo world model $\mathbf{f}$; maximum iterations $M$. \\

Initialize mean $\boldsymbol{\nu}_1$ and covariance $\boldsymbol{\Sigma}_1$ of the Gaussian over joint-velocity sequences $\dot{\boldsymbol{\theta}}_{1:H}$. \\

\For{$m=1$ \KwTo $M$}{
    \vspace{0.1cm}
    Initialize $CostList \gets [\,]$. \\

    \vspace{0.1cm}
    Draw a batch of $n$ velocity-sequence samples
    $
    \left\{
    \tilde{\dot{\boldsymbol{\theta}}}_{1:H,j}
    \right\}_{j=1}^{n}
    \sim
    \mathcal{N}(\boldsymbol{\nu}_m,\boldsymbol{\Sigma}_m)
    $.
    \\

    \vspace{0.1cm}
    $\forall j$: project via QP,
    $
    \dot{\boldsymbol{\theta}}^{*}_{1:H,j}
    =
    \mathrm{QP}
    \left(
    \tilde{\dot{\boldsymbol{\theta}}}_{1:H,j}
    \right)
    $.
    \tcp*[f]{\textcolor{blue}{Projection onto the feasible set enforcing position, velocity, acceleration, and jerk constraints.}} \\

    \vspace{0.1cm}
    Simulate $n$ rollouts in parallel:
    \[
    \left(
    \boldsymbol{\theta}_{k+1,j},
    \mathbf{x}_{k+1,j}
    \right)
    =
    \mathbf{f}
    \left(
    \boldsymbol{\theta}_{k,j},
    \dot{\boldsymbol{\theta}}^{*}_{k,j},
    \mathbf{x}_{k,j}
    \right).
    \]
    \tcp*[f]{\textcolor{blue}{Parallel forward simulation yielding joint trajectories and arm--environment state trajectories.}} \\

    \vspace{0.1cm}
    Evaluate cost components
    $
    \overline{c}_i
    \left(
    \boldsymbol{\theta}_{k,j},
    \dot{\boldsymbol{\theta}}^{*}_{k,j},
    \mathbf{x}_{k,j}
    \right)
    $
    for $k=1,\ldots,H$. \\

    \vspace{0.1cm}
    Aggregate total rollout cost
    \[
    c_j
    =
    \sum_{k=1}^{H}
    \sum_i
    w_i\,
    \overline{c}_i
    \left(
    \boldsymbol{\theta}_{k,j},
    \dot{\boldsymbol{\theta}}^{*}_{k,j},
    \mathbf{x}_{k,j}
    \right),
    \]
    and append to $CostList$. \\

    \vspace{0.1cm}
    Compute importance weights
    $
    \omega_j
    =
    \dfrac{\exp(-c_j/\tau)}
    {\sum_{i=1}^{n}\exp(-c_i/\tau)}
    $.
    \tcp*[f]{\textcolor{blue}{Soft weighting of samples: lower cost $\Rightarrow$ higher weight.}} \\

    \vspace{0.1cm}
    Update mean:
    $
    \boldsymbol{\nu}_{m+1}
    =
    \sum_{j=1}^{n}
    \omega_j\,
    \dot{\boldsymbol{\theta}}^{*}_{1:H,j}
    $.
    \tcp*[f]{\textcolor{blue}{Importance-weighted average in place of elite selection.}} \\

    \vspace{0.1cm}
    Update covariance:
    \[
    \boldsymbol{\Sigma}_{m+1}
    =
    \mathrm{diag}
    \left(
    \sum_{j=1}^{n}
    \omega_j
    \left(
    \dot{\boldsymbol{\theta}}^{*}_{1:H,j}
    -
    \boldsymbol{\nu}_{m+1}
    \right)^{\odot 2}
    \right)
    +
    \epsilon\,\mathbf{I}.
    \]
    \tcp*[f]{\textcolor{blue}{Adaptive, regularized covariance update from weighted samples.}} \\
}

\Return Joint-velocity sequence
$\dot{\boldsymbol{\theta}}^{*}_{1:H}$
corresponding to the lowest-cost sample, together with
$(\boldsymbol{\nu}_M,\boldsymbol{\Sigma}_M)$.

\normalsize
\end{tightalgorithm}

\subsection{Sampling-Based Optimization}

\noindent
Alg.~\ref{algo_1} presents our sampling-based optimizer for solving \eqref{cost}--\eqref{bounds}. The optimizer combines an MPPI~\cite{williams2017model}-style importance-weighted mean update with a weighted diagonal covariance adaptation inspired by the Cross Entropy Method (CEM)~\cite{baseline_cem}. The process begins by initializing a Gaussian distribution over horizon-length joint-velocity sequences
$\dot{\boldsymbol{\theta}}_{1:H}$,
defined by a mean $\boldsymbol{\nu}_m$ and covariance $\boldsymbol{\Sigma}_m$ at iteration $m=1$. This distribution represents the initial search space of candidate control sequences. Within a main loop of $M$ iterations, the optimizer systematically refines this distribution to concentrate around high-quality solutions.

In each iteration, a batch of $n$ candidate velocity sequences
\[
\left\{
\tilde{\dot{\boldsymbol{\theta}}}_{1:H,j}
\right\}_{j=1}^{n}
\]
is drawn from the current distribution. Since these raw samples are typically noisy and may violate the kinematic limits in \eqref{bounds}--\eqref{boundary}, we project each sample onto the feasible set by solving the following quadratic program:
\begin{subequations}
\begin{align}
    \dot{\boldsymbol{\theta}}^{*}_{1:H}
    &=
    \arg\min_{\dot{\boldsymbol{\theta}}_{1:H}}
    \;
    \frac{1}{2}
    \left\|
    \dot{\boldsymbol{\theta}}_{1:H}
    -
    \tilde{\dot{\boldsymbol{\theta}}}_{1:H}
    \right\|_2^2,
    \label{proj_cost}\\
    \text{s.t.}\quad
    &
    \boldsymbol{\theta}^{(r)}_{\min}
    \leq
    \boldsymbol{\theta}^{(r)}_k
    \leq
    \boldsymbol{\theta}^{(r)}_{\max},
    \qquad
    r\in\{0,1,2,3\},
    \quad
    k=1,\ldots,H,
    \label{proj_bounds}\\
    &
    \boldsymbol{\theta}_0
    =
    \boldsymbol{\theta}_{\mathrm{init}},
    \qquad
    \dot{\boldsymbol{\theta}}_0
    =
    \dot{\boldsymbol{\theta}}_{\mathrm{init}}.
    \label{proj_boundary}
\end{align}
\end{subequations}
Positions, accelerations, and jerks are obtained from
$\dot{\boldsymbol{\theta}}_{1:H}$
through discrete-time integration and finite differencing, so that the bounds in \eqref{proj_bounds} jointly enforce position, velocity, acceleration, and jerk limits. The boundary conditions in \eqref{proj_boundary} initialize the projected trajectory at the current robot configuration and ensure consistency with the current joint velocity at the MPC replanning boundary. The QP is solved in parallel across all $n$ samples on the GPU.

With the projected velocity sequences
\[
\left\{
\dot{\boldsymbol{\theta}}^{*}_{1:H,j}
\right\}_{j=1}^{n}
\]
in hand, the algorithm leverages the MuJoCo physics simulator to perform parallel rollouts, predicting the resulting manipulator and environment trajectories
$
(\boldsymbol{\theta}_{k,j},\mathbf{x}_{k,j})
$
over the planning horizon. Each rollout is evaluated using the total trajectory cost
\begin{align}
    c_j
    =
    \sum_{k=1}^{H}
    c\!\left(
    \boldsymbol{\theta}_{k,j},
    \dot{\boldsymbol{\theta}}^{*}_{k,j},
    \mathbf{x}_{k,j}
    \right)
    =
    \sum_{k=1}^{H}
    \sum_i
    w_i\,
    \overline{c}_i
    \left(
    \boldsymbol{\theta}_{k,j},
    \dot{\boldsymbol{\theta}}^{*}_{k,j},
    \mathbf{x}_{k,j}
    \right).
    \label{rollout_cost}
\end{align}
Following a soft weighting scheme, each sample is assigned an importance weight $\omega_j$ through a temperature-scaled exponential of its total cost. These weights drive a moment-matching update of the Gaussian over velocity sequences, in which the mean and covariance are computed as weighted statistics of the projected samples:
\begin{subequations}
\begin{align}
    \omega_j
    &=
    \frac{
    \exp(-c_j/\tau)
    }{
    \sum_{i=1}^{n}
    \exp(-c_i/\tau)
    },
    \label{weight_update}\\
    \boldsymbol{\nu}_{m+1}
    &=
    \sum_{j=1}^{n}
    \omega_j\,
    \dot{\boldsymbol{\theta}}^{*}_{1:H,j},
    \label{mean_update_new}\\
    \boldsymbol{\Sigma}_{m+1}
    &=
    \mathrm{diag}
    \left(
    \sum_{j=1}^{n}
    \omega_j
    \left(
    \dot{\boldsymbol{\theta}}^{*}_{1:H,j}
    -
    \boldsymbol{\nu}_{m+1}
    \right)^{\odot 2}
    \right)
    +
    \epsilon\,\mathbf{I}.
    \label{cov_update_new}
\end{align}
\end{subequations}
Here, $(\cdot)^{\odot 2}$ denotes element-wise squaring, and $\epsilon\,\mathbf{I}$ is a small regularization term that prevents distribution collapse. The mean update in \eqref{mean_update_new} follows the classical MPPI formulation~\cite{williams2017model}, while the covariance update in \eqref{cov_update_new} is a weighted diagonal approximation inspired by the CEM literature~\cite{baseline_cem}.

This cycle of sampling, projection, simulation, evaluation, and distribution update is repeated iteratively, progressively concentrating the search distribution around high-quality solutions. Upon termination, the algorithm returns the joint-velocity sequence
$\dot{\boldsymbol{\theta}}^{*}_{1:H}$
corresponding to the lowest-cost sample, together with the final distribution parameters
$(\boldsymbol{\nu}_M,\boldsymbol{\Sigma}_M)$.
In practice, we execute the first few actions of this velocity sequence and re-plan in a receding-horizon manner, warm-starting the next optimization cycle with
$(\boldsymbol{\nu}_M,\boldsymbol{\Sigma}_M)$.

\subsection{MPC Planner}\label{subsec:MPC Planner}

\begin{figure*}
    \centering
    \includegraphics[width=\linewidth]{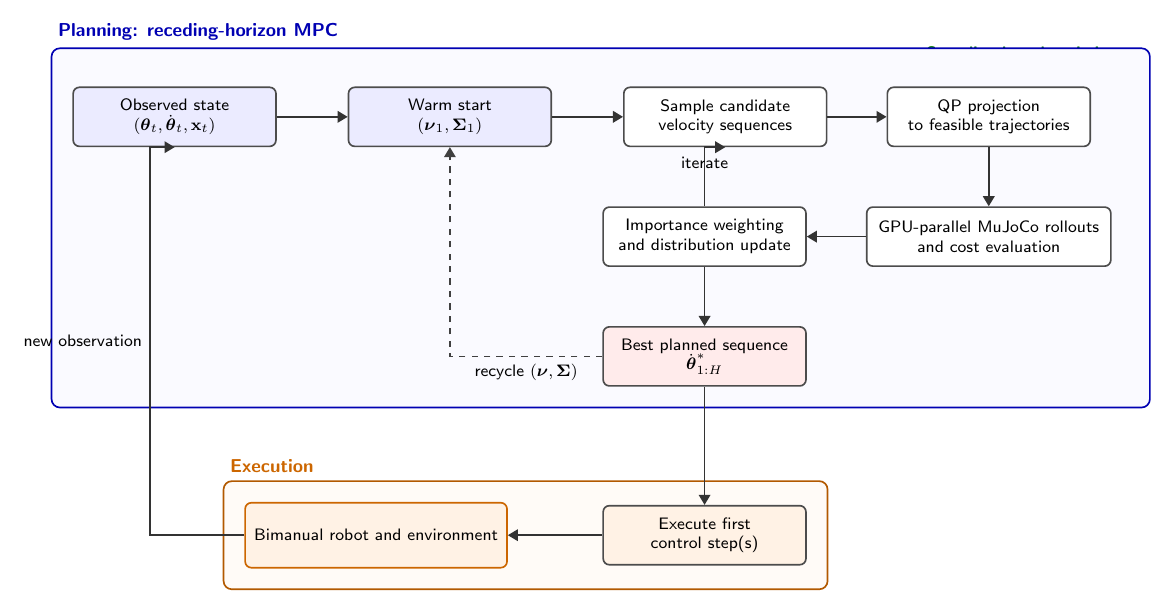}
    \caption{System architecture of our proposed framework. The outer Model Predictive Control (MPC) loop handles real-time state feedback ($\boldsymbol{\theta}, \dot{\boldsymbol{\theta}}, \mathbf{x}$) and warm-starts the distribution parameters. The inner sampling-based optimizer (Alg. 1) enforces constraints via QP projection and leverages highly parallelized MJX rollouts to evaluate task-specific costs.}
    \label{fig:mpc_planner}
\end{figure*}

\noindent
Our MPC planner framework is shown in Fig.~\ref{fig:mpc_planner}. In the proposed architecture, the sampling-based optimizer of Alg.~\ref{algo_1} is embedded within a receding-horizon Model Predictive Control (MPC) loop. At each planning step, the optimizer produces a feasible joint-velocity sequence over a finite horizon $H$. Executing this entire sequence open-loop would be sensitive to simulator-to-reality mismatch, unmodeled contact effects, and changes in the environment, such as object motion or obstacle displacement.

To obtain robust and reactive bimanual behavior, we re-plan at every discrete control step $t$. The planner first observes the current robot and environment state, given by $(\boldsymbol{\theta}_t,\dot{\boldsymbol{\theta}}_t,\mathbf{x}_t)$, and uses this information to initialize the optimization problem. Alg.~\ref{algo_1} is then invoked to compute an optimized joint-velocity sequence over the predictive horizon. Rather than executing the full sequence, only the first few controls are applied to the physical or simulated system, after which a new observation is acquired and the optimization is repeated.

After execution, the prediction horizon is shifted forward. To improve sample efficiency and reduce the number of iterations needed at the next MPC step, we warm-start the optimizer using the distribution obtained from the previous planning cycle. Specifically, the final distribution parameters $(\boldsymbol{\nu}_M,\boldsymbol{\Sigma}_M)$ returned at control step $t$ are time-shifted and recycled to initialize $(\boldsymbol{\nu}_1,\boldsymbol{\Sigma}_1)$ at step $t+1$. This warm-starting mechanism preserves useful trajectory information across consecutive MPC updates while retaining the ability to adapt online to new state feedback.

\subsection{Role of QP}\label{subsec:Role of QP}
 
\noindent The Quadratic Program (QP) in \eqref{proj_cost}--\eqref{proj_boundary} is central to the performance of our sampling-based optimizer. Beyond enforcing kinematic feasibility, it provides two structural advantages that fundamentally improve the behavior of the MPC pipeline: it \emph{decouples exploration from motion smoothness}, and it \emph{stabilizes warm-starting across MPC iterations}. We elaborate on each of these below, followed by an empirical comparison against alternative smoothing strategies.
 
\subsubsection{Decoupling Exploration and Smoothness}\label{subsubsec:decoupling}
 
In standard sampling-based optimizers such as MPPI~\cite{williams2017model} and CEM~\cite{baseline_cem}, the covariance of the sampling distribution simultaneously governs two competing objectives: \emph{exploration} of the action space and \emph{smoothness} of the resulting trajectories. A large covariance promotes wider exploration but produces noisy, jerky samples whose finite-difference derivatives can violate acceleration and jerk limits by orders of magnitude. Conversely, a small covariance yields smooth samples but restricts exploration to a narrow neighborhood of the current mean, risking convergence to poor local optima---a particularly acute problem in contact-rich bimanual tasks where the cost landscape is highly non-convex.
 
The QP projection eliminates this tension. By allowing the user to specify an arbitrarily large sampling covariance and then projecting each sample onto the feasible set defined by \eqref{proj_bounds}--\eqref{proj_boundary}, the optimizer can explore aggressively while guaranteeing that every evaluated trajectory satisfies strict bounds on position, velocity, acceleration, and jerk. Crucially, the projection is \emph{minimal} in the $\ell_2$ sense: it finds the closest feasible trajectory to each raw sample, thereby preserving as much of the original exploratory perturbation as possible. The result is a set of samples that comprehensively span the \emph{reachable} portion of the action space---fanning out to saturate at the kinematic limits---without any of them inducing oscillatory or dynamically inconsistent motions.
 
\subsubsection{Stabilizing Warm-Start Across MPC Iterations}
 
The receding-horizon MPC loop (Section~\ref{subsec:MPC Planner}) relies on warm-starting successive planning calls with the distribution parameters $(\boldsymbol{\nu}_M, \boldsymbol{\Sigma}_M)$ from the previous step. The effectiveness of this warm-start depends critically on how much the optimal trajectory changes between two consecutive MPC iterations. If the commanded control profile can oscillate sharply from one step to the next---as commonly occurs in unconstrained MPPI---the warm-start distribution becomes a poor prior, degrading sample efficiency and destabilizing the overall pipeline.

The derivative bounds enforced by the QP directly mitigate this problem. Because every projected sample satisfies bounded velocity, acceleration, and jerk, the planned trajectories are inherently smooth and cannot exhibit the sharp temporal variations that cause consecutive MPC solutions to diverge. In particular, the jerk bound limits how rapidly the acceleration profile can change, which in turn limits how rapidly the velocity profile can evolve. This cascading effect ensures that trajectories optimized at successive MPC steps remain close to each other, keeping the warm-start distribution well-aligned with the new optimum and yielding faster convergence of Alg.~\ref{algo_1}. This observation is consistent with the findings of~\cite{andrejev2025pi}, where bounded control derivatives were shown to prevent the inter-iteration oscillations that degrade MPPI warm-starting.
 
\subsubsection{Comparison With Frequency-Domain Smoothing}
 
Recent works such as LP-MPPI~\cite{kicki2025low} address the smoothness problem from the frequency domain by applying a low-pass filter (LPF) to the sampled perturbations before rollout. While this attenuates high-frequency energy and introduces temporal correlation between successive controls, it has two fundamental limitations compared to the QP projection.
 
First, a low-pass filter \emph{cannot enforce strict bounds} on control magnitudes or their derivatives. It shapes the power spectral density of the perturbations but provides no guarantee that any individual sample satisfies $\boldsymbol{\theta}^{(r)}_{\min} \leq \boldsymbol{\theta}_k^{(r)} \leq \boldsymbol{\theta}^{(r)}_{\max}$. A filtered sample with only low-frequency content can still violate derivative bounds if its amplitude is sufficiently large. In contrast, the QP produces certificates of feasibility for every sample.

Second, low-pass filtering provides no mechanism to pin the initial velocity to the current robot state. The QP enforces $\dot{\boldsymbol{\theta}}_1^* = \dot{\boldsymbol{\theta}}_{\mathrm{init}}$ as an equality constraint \eqref{proj_boundary}, ensuring that every projected sample starts from the current commanded velocity. A stateless filter applied to perturbations has no analogue for this: even if filtered samples are individually smooth, the first commanded velocity can differ arbitrarily from the previous MPC step's output, introducing discontinuities at the replanning boundary.
 
\subsubsection{Empirical Validation}
 
The arguments of the preceding subsections make two testable predictions. First, the QP projection should permit substantially larger sampling covariances than either unconstrained (baseline) or frequency-domain (LPF) sampling, without violating any kinematic bound. Second, this enlarged covariance should translate into qualitatively different sampling behavior: QP-projected trajectories should span the full kinematically reachable set, whereas baseline and LPF trajectories---constrained to use conservative covariances to avoid violations---should remain clustered in a narrow region around the mean. We design the following experiment to test both predictions on a single-joint trajectory planning problem with representative kinematic limits: velocity $\dot{\theta}_{\max} = \pm 1\;\mathrm{rad/s}$, acceleration $\ddot{\theta}_{\max} = \pm 2\;\mathrm{rad/s}^2$, and jerk $\dddot{\theta}_{\max} = \pm 4\;\mathrm{rad/s}^3$, and an initial velocity condition $\dot{\theta}_1 = \dot{\theta}_{\mathrm{init}}$. We emphasize that the following single-joint experiment is a simplified diagnostic study, not a manipulation benchmark. Its purpose is to isolate the sampling behavior of the compared methods under identical kinematic bounds, independent of task costs, contact dynamics, or bimanual coordination. This toy example therefore serves only to illustrate how QP projection changes the distribution of feasible samples relative to unfiltered and low-pass-filtered alternatives.

\parletter{Objective}
For each sampling strategy---baseline (uncorrelated Gaussian), LPF (second-order Butterworth filtered Gaussian), and QP projection (ours)---we seek to determine the \emph{maximum effective standard deviation} $\sigma_{\mathrm{eff}}$ of the sampling distribution under which the method can still guarantee strict satisfaction of all kinematic bounds across the entire planning horizon. This quantity directly measures the exploration budget available to the optimizer: a larger $\sigma_{\mathrm{eff}}$ means the method can search over a wider region of the action space while remaining dynamically feasible.
 
\parletter{Methodology}
For the baseline and LPF methods, we employ a Monte Carlo binary search to identify $\sigma_{\mathrm{eff}}$. At each iteration of the binary search, we fix a candidate standard deviation $\sigma$ and draw $N = 200{,}000$ independent trajectory samples from $\mathcal{N}(0,\, \sigma^2 \mathbf{I})$ over the planning horizon. For the baseline, these samples are used directly as joint velocity sequences; for LPF, each sample is passed through a second-order Butterworth filter with a cutoff frequency tuned to the system bandwidth before use. From each velocity sequence, we compute the corresponding position (via cumulative integration), acceleration (via first-order finite differencing of velocity), and jerk (via second-order finite differencing) profiles, and check whether prescribed bounds are satisfied at every time step. The violation rate is defined as the fraction of trajectories in which \emph{any} bound is violated at \emph{any} time step. We declare a candidate $\sigma$ admissible if the violation rate falls below $0.3\%$, corresponding to a $99.7\%$ confidence level---the three-sigma threshold. The binary search terminates when the interval width falls below a tolerance of $10^{-4}$, at which point the largest admissible $\sigma$ is reported as $\sigma_{\mathrm{eff}}$.
 
For the QP method, the concept of $\sigma_{\mathrm{eff}}$ takes on a different meaning. Since the projection maps \emph{every} sample onto the feasible set by construction, there is no covariance at which violations occur. Instead, we characterize its effective covariance as follows: we draw $200{,}000$ samples from $\mathcal{N}(0,\, \sigma^2 \mathbf{I})$ with a large $\sigma$, project each through the QP \eqref{proj_cost}--\eqref{proj_boundary} (enforcing velocity, acceleration, and jerk bounds along with the initial velocity condition), and then compute the empirical standard deviation of the \emph{projected} velocity sequences. This post-projection standard deviation, denoted $\sigma_{\mathrm{eff}}^{\mathrm{QP}}$, reflects the actual spread of the feasible samples that enter the rollout and cost evaluation stages of Alg.~\ref{algo_1}. It represents the effective exploration radius in the space of \emph{feasible} trajectories.
 
\begin{figure}
    \centering
    \includegraphics[width=\linewidth]{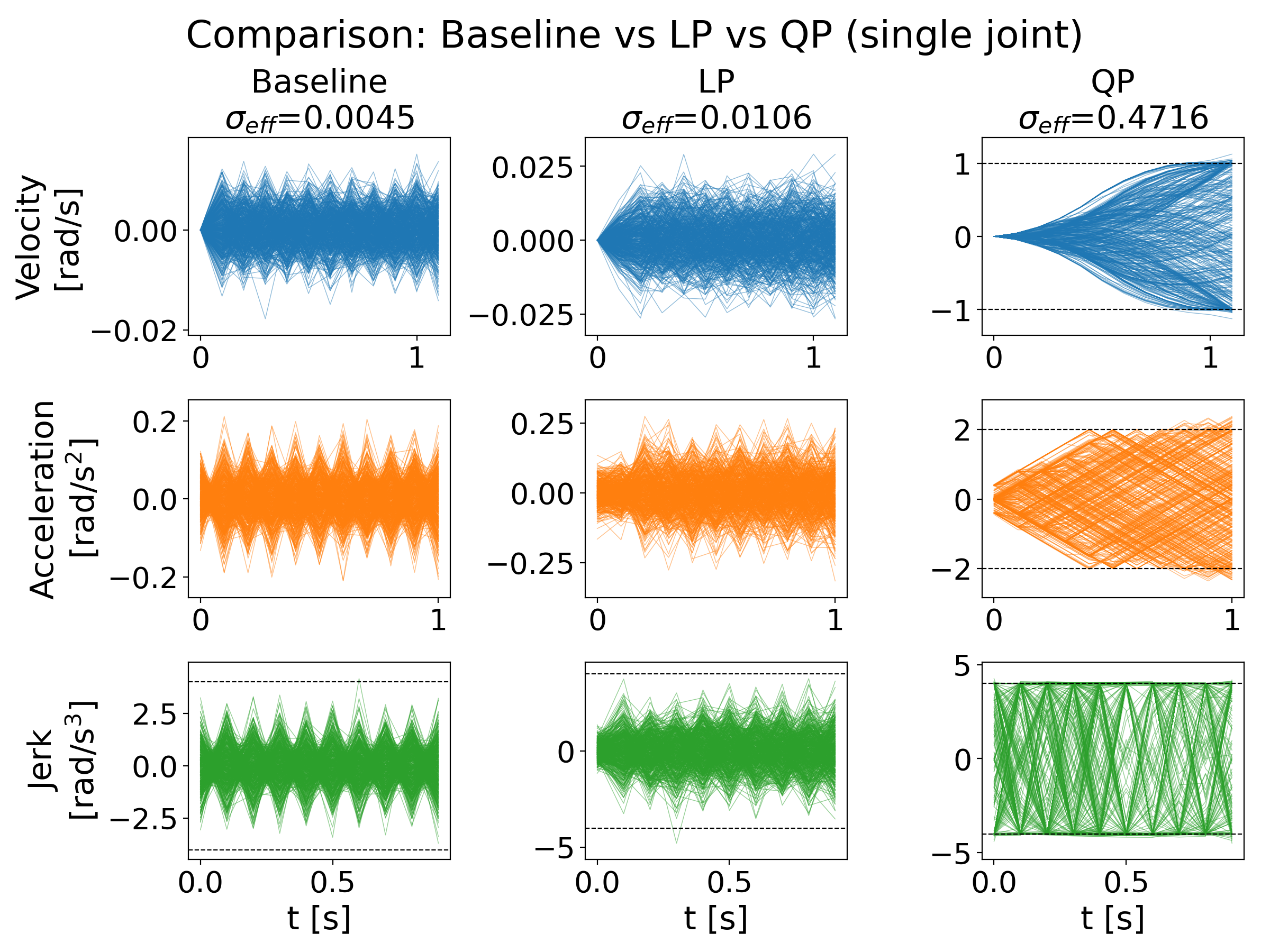}
    \caption{Comparison of sampling behaviors for a single joint. The QP-projected samples (right) share a common initial velocity $\dot{\theta}_1 = \dot{\theta}_{\mathrm{init}}$ and fan out to cover the full feasible envelope defined by velocity, acceleration, and jerk bounds. The baseline (left) and LPF (middle) are restricted to conservative covariances, yielding negligible coverage of the reachable set.}
    \label{fig:comparison_baseline_lp_qp}
\end{figure}

\parletter{Results}
The results are summarized in Fig.~\ref{fig:comparison_baseline_lp_qp} and reveal stark differences in the exploration capacity of the three methods.
 
\noindent\textit{Baseline (uncorrelated white noise):}
The baseline yields the most restrictive effective covariance, $\sigma_{\mathrm{eff}} = 0.0045$. This extreme conservatism arises because independent samples at successive time steps produce velocity sequences that are essentially discrete white noise. Finite differencing of white noise amplifies spectral energy quadratically with the derivative order: if $\dot{\theta}_k$ has variance $\sigma^2$, then $\ddot{\theta}_k \approx (\dot{\theta}_{k+1} - \dot{\theta}_k)/\Delta t$ has variance $2\sigma^2/\Delta t^2$, and $\dddot{\theta}_k$ has variance $6\sigma^2/\Delta t^4$. Consequently, the jerk bound is violated long before the velocity or acceleration bounds become active, and the entire covariance budget is consumed by the highest-order constraint. The resulting trajectories, shown in the left panel of Fig.~\ref{fig:comparison_baseline_lp_qp}, are tightly clustered around zero with negligible spatial coverage.
 
\noindent\textit{Low-pass filter (LPF):}
Applying a second-order Butterworth filter introduces temporal correlation between successive velocity samples, attenuating the high-frequency energy that drives jerk violations. This yields a modest improvement to $\sigma_{\mathrm{eff}} = 0.0106$---a factor of $2.4\times$ over the baseline. The improvement is limited for two reasons. First, the filter \emph{attenuates} high frequencies but does not \emph{eliminate} them; residual energy above the cutoff still contributes to derivative magnitudes, and with $200{,}000$ samples, even rare large excursions trigger the $0.3\%$ violation threshold. Second, the filter is amplitude-agnostic: it scales all frequency components of a sample by fixed gain factors, so doubling $\sigma$ doubles the amplitude of the filtered output and consequently doubles all derivative magnitudes. There is therefore no decoupling between sampling variance and derivative compliance. The middle panel of Fig.~\ref{fig:comparison_baseline_lp_qp} confirms that LPF trajectories, while slightly more spread than the baseline, remain confined to a small fraction of the feasible region.
 
\noindent\textit{QP projection (ours):}
The QP-projected samples exhibit a qualitatively different behavior. With $\sigma_{\mathrm{eff}}^{\mathrm{QP}} = 0.4716$, the effective exploration radius is approximately $100\times$ that of the baseline and $45\times$ that of LPF. The right panel of Fig.~\ref{fig:comparison_baseline_lp_qp} illustrates this: projected velocity trajectories fan out to fill the entire feasible envelope, with individual trajectories reaching and saturating at the velocity bounds ($\pm 1\;\mathrm{rad/s}$), acceleration bounds ($\pm 2\;\mathrm{rad/s}^2$), and jerk bounds ($\pm 4\;\mathrm{rad/s}^3$) without exceeding them. This saturation behavior is a direct consequence of the minimal $\ell_2$ projection: samples drawn far outside the feasible set are mapped to the nearest boundary point, naturally populating the extremes of the reachable space. Importantly, all projected samples share the same initial velocity $\dot{\theta}_1 = \dot{\theta}_{\mathrm{init}}$ (visible as the common starting point in the right panel of Fig.~\ref{fig:comparison_baseline_lp_qp}), after which they diverge to cover the full kinematically admissible region---precisely the property needed for effective exploration in high-dimensional, non-convex optimization landscapes.
 
\parletter{Implications for Task Performance}
The two-orders-of-magnitude gap in effective covariance has direct consequences for the downstream manipulation tasks. In the contact-rich bimanual setting, the optimizer must discover coordinated arm trajectories that navigate narrow passages between obstacles, establish and maintain frictional contact with objects, and transport payloads along collision-free paths. These behaviors correspond to small, isolated regions in the joint-velocity space, and finding them requires dense coverage of the feasible set. The baseline and LPF methods, restricted to $\sigma_{\mathrm{eff}} \leq 0.011$, sample almost exclusively near the current mean and are unlikely to discover such solutions without many MPC iterations. On the other hand, if we disregard the velocity, acceleration and jerk bounds for the baseline and LPF, then the trajectory oscillations between subsequent MPC steps degrade the effectiveness of warm-starting and consequently performance. The QP projection, by contrast, generates samples that span the full reachable set at every call to Alg.~\ref{algo_1}, substantially increasing the probability of discovering high-quality trajectories within a fixed sample budget. This advantage is borne out quantitatively in the task-level comparisons of Section~\ref{subsec:Baseline comparison}, where the QP-based method achieves consistently higher success rates than LPF and baseline across all four bimanual tasks.

\subsection{Custom QP Solver}\label{subsec:custom_qp_solver}

\noindent In this section, we derive a computationally efficient, GPU-parallelizable solver for the projection QP \eqref{proj_cost}--\eqref{proj_boundary}, adapting the ADMM/Bregman-splitting scheme of~\cite{andrejev2025pi,ghadimi2014optimal} to the bimanual joint-velocity setting. Throughout the derivation, $\boldsymbol{\theta}_k \in \mathbb{R}^{N_{\mathrm{dof}}}$ denotes the combined joint configuration of both manipulators at planning step $k$, where $N_{\mathrm{dof}}$ is the total number of joints.

Although the derivation below is written in stacked vector form over all joints and horizon steps, the resulting QP remains \emph{separable across joints}. This follows from the fact that both the projection objective and the derivative constraints act component-wise on each joint coordinate. In implementation, we exploit this structure by solving $N_{\mathrm{dof}}$ independent horizon-wise QPs in parallel on the GPU.

\subsubsection{From Joint-Velocity Sequences to Derivative Bounds}

Let
\begin{align}
    \tilde{\dot{\boldsymbol{\Theta}}}
    =
    \operatorname{col}
    \left(
    \tilde{\dot{\boldsymbol{\theta}}}_1,\,
    \tilde{\dot{\boldsymbol{\theta}}}_2,\,
    \ldots,\,
    \tilde{\dot{\boldsymbol{\theta}}}_H
    \right)
    \in \mathbb{R}^{H N_{\mathrm{dof}}}
\end{align}
denote a sampled joint-velocity trajectory over the planning horizon, where
$\tilde{\dot{\boldsymbol{\theta}}}_k \in \mathbb{R}^{N_{\mathrm{dof}}}$ is the sampled joint-velocity vector at step $k$. Similarly, let
\begin{align}
    \dot{\boldsymbol{\Theta}}^{*}
    =
    \operatorname{col}
    \left(
    \dot{\boldsymbol{\theta}}^{*}_1,\,
    \dot{\boldsymbol{\theta}}^{*}_2,\,
    \ldots,\,
    \dot{\boldsymbol{\theta}}^{*}_H
    \right)
    \in \mathbb{R}^{H N_{\mathrm{dof}}}
\end{align}
denote its projected feasible counterpart. The projection QP seeks the closest feasible joint-velocity sequence:
\begin{align}
    \min_{\dot{\boldsymbol{\Theta}}^{*}}
    \;\;
    \frac{1}{2}
    \left\|
    \dot{\boldsymbol{\Theta}}^{*}
    -
    \tilde{\dot{\boldsymbol{\Theta}}}
    \right\|_2^2.
    \label{eq:qp_obj_scalar}
\end{align}

\noindent The projected trajectory must satisfy bounds on joint position, velocity, acceleration, and jerk. These quantities can all be expressed as affine functions of $\dot{\boldsymbol{\Theta}}^{*}$.

\paragraph{Velocity bounds ($r=1$)}
The velocity constraints act directly on the decision variable:
\begin{align}
    \mathbf{1}_H \otimes \dot{\boldsymbol{\theta}}_{\min}
    \leq
    \dot{\boldsymbol{\Theta}}^{*}
    \leq
    \mathbf{1}_H \otimes \dot{\boldsymbol{\theta}}_{\max},
    \label{eq:vel_bounds}
\end{align}
where
$\dot{\boldsymbol{\theta}}_{\min}, \dot{\boldsymbol{\theta}}_{\max}
\in \mathbb{R}^{N_{\mathrm{dof}}}$
are the joint-wise velocity limits, $\mathbf{1}_H$ is a vector of ones, and $\otimes$ denotes the Kronecker product.
Throughout, $\mathbf{1}_H \in \mathbb{R}^H$ denotes the all-ones vector, so
$\mathbf{1}_H \otimes \boldsymbol{v} \in \mathbb{R}^{H N_{\mathrm{dof}}}$
denotes the $H$-fold vertical tiling of any vector
$\boldsymbol{v} \in \mathbb{R}^{N_{\mathrm{dof}}}$.

\paragraph{Position bounds ($r=0$)}
Joint positions are obtained by forward Euler integration from the current robot configuration $\boldsymbol{\theta}_{\mathrm{init}}$:
\begin{align}
    \boldsymbol{\theta}^{*}_k
    =
    \boldsymbol{\theta}_{\mathrm{init}}
    +
    \Delta t
    \sum_{i=1}^{k}
    \dot{\boldsymbol{\theta}}^{*}_i,
    \qquad
    k=1,\ldots,H.
\end{align}
Define the stacked joint-position trajectory
\begin{align}
    \boldsymbol{\Theta}^{*}
    =
    \operatorname{col}
    \left(
    \boldsymbol{\theta}^{*}_1,\,
    \boldsymbol{\theta}^{*}_2,\,
    \ldots,\,
    \boldsymbol{\theta}^{*}_H
    \right)
    \in \mathbb{R}^{H N_{\mathrm{dof}}}.
\end{align}
Then,
\begin{align}
    \boldsymbol{\Theta}^{*}
    =
    \mathbf{1}_H \otimes \boldsymbol{\theta}_{\mathrm{init}}
    +
    \Delta t\,
    \mathbf{L}_{\mathrm{dof}}
    \dot{\boldsymbol{\Theta}}^{*},
    \label{eq:pos_stack}
\end{align}
where
\begin{align}
    \mathbf{L}_{\mathrm{dof}}
    =
    \mathbf{L} \otimes \mathbf{I}_{N_{\mathrm{dof}}},
\end{align}
$\mathbf{I}_{N_{\mathrm{dof}}}\in
\mathbb{R}^{N_{\mathrm{dof}}\times N_{\mathrm{dof}}}$
is the identity matrix in joint space, and
$\mathbf{L}\in\mathbb{R}^{H\times H}$ is the lower-triangular cumulative-sum matrix
\begin{align}
    \mathbf{L}
    =
    \begin{bmatrix}
        1 & 0 & \cdots & 0 \\
        1 & 1 & \ddots & \vdots \\
        \vdots & \ddots & \ddots & 0 \\
        1 & \cdots & 1 & 1
    \end{bmatrix}.
\end{align}
The position bounds therefore become
\begin{align}
    \mathbf{1}_H \otimes \boldsymbol{\theta}_{\min}
    \leq
    \mathbf{1}_H \otimes \boldsymbol{\theta}_{\mathrm{init}}
    +
    \Delta t\,\mathbf{L}_{\mathrm{dof}}
    \dot{\boldsymbol{\Theta}}^{*}
    \leq
    \mathbf{1}_H \otimes \boldsymbol{\theta}_{\max}.
    \label{eq:pos_bounds}
\end{align}
The initial position condition
$\boldsymbol{\theta}_0=\boldsymbol{\theta}_{\mathrm{init}}$
is thus incorporated directly through the integration model.

\paragraph{Acceleration bounds ($r=2$)}
Accelerations are approximated through first-order backward differencing of consecutive joint-velocity vectors. The velocity preceding the first planning step is taken as the current MPC boundary value $\dot{\boldsymbol{\theta}}_{\mathrm{init}}$:
\begin{align}
    \ddot{\boldsymbol{\theta}}^{*}_k
    =
    \frac{
    \dot{\boldsymbol{\theta}}^{*}_k
    -
    \dot{\boldsymbol{\theta}}^{*}_{k-1}}
    {\Delta t},
    \qquad
    k=1,\ldots,H,
\end{align}
with
\begin{align}
    \dot{\boldsymbol{\theta}}^{*}_0
    \triangleq
    \dot{\boldsymbol{\theta}}_{\mathrm{init}}.
\end{align}
Define the stacked acceleration trajectory
\begin{align}
    \ddot{\boldsymbol{\Theta}}^{*}
    =
    \operatorname{col}
    \left(
    \ddot{\boldsymbol{\theta}}^{*}_1,\,
    \ddot{\boldsymbol{\theta}}^{*}_2,\,
    \ldots,\,
    \ddot{\boldsymbol{\theta}}^{*}_H
    \right)
    \in \mathbb{R}^{H N_{\mathrm{dof}}}.
\end{align}
Then,
\begin{align}
    \ddot{\boldsymbol{\Theta}}^{*}
    =
    \frac{1}{\Delta t}
    \mathbf{D}_{1,\mathrm{dof}}
    \dot{\boldsymbol{\Theta}}^{*}
    +
    \mathbf{b}_{\mathrm{acc}},
    \label{eq:acc_stack}
\end{align}
where
\begin{align}
    \mathbf{D}_{1,\mathrm{dof}}
    =
    \mathbf{D}_1 \otimes \mathbf{I}_{N_{\mathrm{dof}}},
\end{align}
with
\begin{align}
    \mathbf{D}_1
    =
    \begin{bmatrix}
        1 & 0 & \cdots & 0 \\
        -1 & 1 & \ddots & \vdots \\
        0 & \ddots & \ddots & 0 \\
        0 & \cdots & -1 & 1
    \end{bmatrix},
\end{align}
and
\begin{align}
    \mathbf{b}_{\mathrm{acc}}
    =
    \frac{1}{\Delta t}
    \operatorname{col}
    \left(
    -\dot{\boldsymbol{\theta}}_{\mathrm{init}},
    \mathbf{0},
    \ldots,
    \mathbf{0}
    \right)
    \in \mathbb{R}^{H N_{\mathrm{dof}}}.
    \label{eq:b_acc}
\end{align}
The acceleration bounds are therefore
\begin{align}
    \mathbf{1}_H \otimes \ddot{\boldsymbol{\theta}}_{\min}
    \leq
    \frac{1}{\Delta t}
    \mathbf{D}_{1,\mathrm{dof}}
    \dot{\boldsymbol{\Theta}}^{*}
    +
    \mathbf{b}_{\mathrm{acc}}
    \leq
    \mathbf{1}_H \otimes \ddot{\boldsymbol{\theta}}_{\max}.
    \label{eq:acc_bounds}
\end{align}

\paragraph{Jerk bounds ($r=3$)}
{Jerks are computed by differencing consecutive acceleration vectors. }:
\begin{align}
    \dddot{\boldsymbol{\theta}}^{*}_k
    =
    \frac{
    \ddot{\boldsymbol{\theta}}^{*}_k
    -
    \ddot{\boldsymbol{\theta}}^{*}_{k-1}}
    {\Delta t},
    \qquad
    k=2,\ldots,H.
\end{align}
Substituting the velocity-difference expression for acceleration gives
\begin{align}
    \dddot{\boldsymbol{\theta}}^{*}_k
    =
    \frac{
    \dot{\boldsymbol{\theta}}^{*}_k
    -
    2\dot{\boldsymbol{\theta}}^{*}_{k-1}
    +
    \dot{\boldsymbol{\theta}}^{*}_{k-2}}
    {\Delta t^2},
    \qquad
    k=2,\ldots,H,
\end{align}
where the first valid jerk term uses
$\dot{\boldsymbol{\theta}}^{*}_0
=
\dot{\boldsymbol{\theta}}_{\mathrm{init}}$.

{Define the stacked jerk trajectory}
\begin{align}
    \dddot{\boldsymbol{\Theta}}^{*}
    =
    \operatorname{col}
    \left(
    \dddot{\boldsymbol{\theta}}^{*}_2,\,
    \dddot{\boldsymbol{\theta}}^{*}_3,\,
    \ldots,\,
    \dddot{\boldsymbol{\theta}}^{*}_H
    \right)
    \in \mathbb{R}^{(H-1)N_{\mathrm{dof}}}.
\end{align}
Then,
\begin{align}
    \dddot{\boldsymbol{\Theta}}^{*}
    =
    \frac{1}{\Delta t^2}
    \mathbf{D}_{2,\mathrm{dof}}
    \dot{\boldsymbol{\Theta}}^{*}
    +
    \mathbf{b}_{\mathrm{jerk}},
    \label{eq:jerk_stack}
\end{align}
where
\begin{align}
    \mathbf{D}_{2,\mathrm{dof}}
    =
    \mathbf{D}_2 \otimes \mathbf{I}_{N_{\mathrm{dof}}},
\end{align}
with
\begin{align}
    \mathbf{D}_2
    =
    \begin{bmatrix}
        -2 & 1 & 0 & \cdots & 0 \\
        1 & -2 & 1 & \ddots & \vdots \\
        0 & \ddots & \ddots & \ddots & 0 \\
        0 & \cdots & 1 & -2 & 1
    \end{bmatrix}
    \in \mathbb{R}^{(H-1)\times H},
    \label{eq:D2_def}
\end{align}
and
\begin{align}
    \mathbf{b}_{\mathrm{jerk}}
    =
    \frac{1}{\Delta t^2}
    \operatorname{col}
    \left(
    \dot{\boldsymbol{\theta}}_{\mathrm{init}},
    \mathbf{0},
    \ldots,
    \mathbf{0}
    \right)
    \in \mathbb{R}^{(H-1)N_{\mathrm{dof}}}.
    \label{eq:b_jerk}
\end{align}
The jerk bounds become
\begin{align}
    \mathbf{1}_{H-1} \otimes \dddot{\boldsymbol{\theta}}_{\min}
    \leq
    \frac{1}{\Delta t^2}
    \mathbf{D}_{2,\mathrm{dof}}
    \dot{\boldsymbol{\Theta}}^{*}
    +
    \mathbf{b}_{\mathrm{jerk}}
    \leq
    \mathbf{1}_{H-1} \otimes \dddot{\boldsymbol{\theta}}_{\max}.
    \label{eq:jerk_bounds}
\end{align}

\subsubsection{Compact QP Formulation}

Stacking the inequality constraints
\eqref{eq:vel_bounds},
\eqref{eq:pos_bounds},
\eqref{eq:acc_bounds},
and
\eqref{eq:jerk_bounds}
into a single system yields
\begin{subequations}\label{eq:compact_qp}
\begin{align}
    \min_{\dot{\boldsymbol{\Theta}}^{*},\,\boldsymbol{\zeta}}
    \;\;
    &
    \frac{1}{2}
    \left\|
    \dot{\boldsymbol{\Theta}}^{*}
    -
    \tilde{\dot{\boldsymbol{\Theta}}}
    \right\|_2^2,
    \label{eq:compact_obj}\\
    \text{s.t.}\;\;
    &
    \mathbf{A}
    \dot{\boldsymbol{\Theta}}^{*}
    =
    \mathbf{b},
    \label{eq:compact_eq}\\
    &
    \mathbf{G}
    \dot{\boldsymbol{\Theta}}^{*}
    -
    \mathbf{h}
    +
    \boldsymbol{\zeta}
    =
    \mathbf{0},
    \label{eq:compact_ineq}\\
    &
    \boldsymbol{\zeta}
    \geq
    \mathbf{0}.
    \label{eq:compact_slack}
\end{align}
\end{subequations}

The inequality matrix
\[
\mathbf{G}
\in
\mathbb{R}^{(8H-2)N_{\mathrm{dof}} \times HN_{\mathrm{dof}}}
\]
and the corresponding bound vector
\[
\mathbf{h}
\in
\mathbb{R}^{(8H-2)N_{\mathrm{dof}}}
\]
are defined as
\begin{align}
    \mathbf{G}
    =
    \begin{bmatrix}
        \mathbf{I}_{HN_{\mathrm{dof}}} \\
        -\mathbf{I}_{HN_{\mathrm{dof}}} \\
        \Delta t\,\mathbf{L}_{\mathrm{dof}} \\
        -\Delta t\,\mathbf{L}_{\mathrm{dof}} \\
        \frac{1}{\Delta t}\mathbf{D}_{1,\mathrm{dof}} \\
        -\frac{1}{\Delta t}\mathbf{D}_{1,\mathrm{dof}} \\
        \frac{1}{\Delta t^2}\mathbf{D}_{2,\mathrm{dof}} \\
        -\frac{1}{\Delta t^2}\mathbf{D}_{2,\mathrm{dof}}
    \end{bmatrix},
\end{align}
where
$\mathbf{I}_{HN_{\mathrm{dof}}}
\in
\mathbb{R}^{HN_{\mathrm{dof}}\times HN_{\mathrm{dof}}}$
is the identity matrix in the stacked trajectory space. The corresponding bound vector is
\begin{align}
    \mathbf{h}
    =
    \begin{bmatrix}
        \mathbf{1}_H \otimes \dot{\boldsymbol{\theta}}_{\max} \\
        -\mathbf{1}_H \otimes \dot{\boldsymbol{\theta}}_{\min} \\
        \mathbf{1}_H \otimes
        \left(
        \boldsymbol{\theta}_{\max}
        -
        \boldsymbol{\theta}_{\mathrm{init}}
        \right) \\
        -\mathbf{1}_H \otimes
        \left(
        \boldsymbol{\theta}_{\min}
        -
        \boldsymbol{\theta}_{\mathrm{init}}
        \right) \\
        \mathbf{1}_H \otimes \ddot{\boldsymbol{\theta}}_{\max}
        -
        \mathbf{b}_{\mathrm{acc}} \\
        -\mathbf{1}_H \otimes \ddot{\boldsymbol{\theta}}_{\min}
        +
        \mathbf{b}_{\mathrm{acc}} \\
        \mathbf{1}_{H-1} \otimes \dddot{\boldsymbol{\theta}}_{\max}
        -
        \mathbf{b}_{\mathrm{jerk}} \\
        -\mathbf{1}_{H-1} \otimes \dddot{\boldsymbol{\theta}}_{\min}
        +
        \mathbf{b}_{\mathrm{jerk}}
    \end{bmatrix}.
    \label{eq:G_h_def}
\end{align}
The inequality
\begin{align}
    \mathbf{G}\dot{\boldsymbol{\Theta}}^{*}
    \leq
    \mathbf{h}
\end{align}
is therefore equivalent to enforcing all position, velocity, acceleration, and jerk bounds simultaneously. The slack variable $\boldsymbol{\zeta}\geq\mathbf{0}$ converts these inequalities into the equality form in~\eqref{eq:compact_ineq}.

The equality constraint
\[
\mathbf{A}\dot{\boldsymbol{\Theta}}^{*}
=
\mathbf{b}
\]
encodes the initial-velocity boundary condition
\begin{align}
    \dot{\boldsymbol{\theta}}^{*}_1
    =
    \dot{\boldsymbol{\theta}}_{\mathrm{init}}.
\end{align}
Specifically,
\begin{align}
    \mathbf{A}
    =
    \begin{bmatrix}
        \mathbf{I}_{N_{\mathrm{dof}}}
        &
        \mathbf{0}
        &
        \cdots
        &
        \mathbf{0}
    \end{bmatrix}
    \in
    \mathbb{R}^{N_{\mathrm{dof}}\times HN_{\mathrm{dof}}},
    \quad
    \mathbf{b}
    =
    \dot{\boldsymbol{\theta}}_{\mathrm{init}}
    \in
    \mathbb{R}^{N_{\mathrm{dof}}}.
    \label{eq:A_b_def}
\end{align}
Thus, the projected velocity trajectory remains continuous with the command applied at the MPC replanning boundary. Acceleration and jerk are constrained through the inequality system, but no separate equality boundary conditions are imposed on them.

\subsubsection{ADMM Solver}

Following~\cite{andrejev2025pi,ghadimi2014optimal}, we solve \eqref{eq:compact_qp} using an ADMM-inspired Bregman-splitting scheme. The method alternates between a projection onto the nonnegative slack space, a constrained quadratic minimization in the primal trajectory space, and a residual-driven update of a primal-space Bregman multiplier. Let
\[
\boldsymbol{\lambda}
\in
\mathbb{R}^{HN_{\mathrm{dof}}}
\]
be a Bregman multiplier that resides in the same space as the primal variable.

The augmented objective is
\begin{align}
    \mathcal{L}
    \left(
    \dot{\boldsymbol{\Theta}}^{*},
    \boldsymbol{\lambda},
    \boldsymbol{\zeta}
    \right)
    =
    &\;
    \frac{1}{2}
    \left\|
    \dot{\boldsymbol{\Theta}}^{*}
    -
    \tilde{\dot{\boldsymbol{\Theta}}}
    \right\|_2^2
    -
    \left\langle
    \boldsymbol{\lambda},
    \dot{\boldsymbol{\Theta}}^{*}
    \right\rangle
    \notag\\
    &+
    \frac{\delta}{2}
    \left\|
    \mathbf{G}
    \dot{\boldsymbol{\Theta}}^{*}
    -
    \mathbf{h}
    +
    \boldsymbol{\zeta}
    \right\|_2^2,
    \label{eq:aug_lag}
\end{align}
where $\delta>0$ is the penalty parameter. The iterations proceed as follows.

\paragraph{Step 1: Slack update}
At iteration $l+1$, the slack variable is updated through projection onto the nonnegative orthant:
\begin{align}
    {}^{l+1}\boldsymbol{\zeta}
    =
    \max
    \left\{
    \mathbf{0},
    \;
    -\mathbf{G}\,
    {}^{l}\dot{\boldsymbol{\Theta}}^{*}
    +
    \mathbf{h}
    \right\}.
    \label{eq:admm_zeta}
\end{align}
This is an element-wise clipping operation and requires no matrix factorization.

\paragraph{Step 2: Primal update}
The projected velocity trajectory is obtained by minimizing
\eqref{eq:aug_lag}
with respect to
$\dot{\boldsymbol{\Theta}}^{*}$
subject to
\[
\mathbf{A}
\dot{\boldsymbol{\Theta}}^{*}
=
\mathbf{b}.
\]
This yields
\begin{align}
    {}^{l+1}\dot{\boldsymbol{\Theta}}^{*}
    =
    \mathbf{M}^{-1}
    \mathbf{r}
    \left(
    \tilde{\dot{\boldsymbol{\Theta}}},
    {}^{l}\boldsymbol{\lambda},
    {}^{l+1}\boldsymbol{\zeta}
    \right),
    \label{eq:admm_primal}
\end{align}
where
\begin{align}
    \mathbf{M}
    =
    \begin{bmatrix}
        \mathbf{I}_{HN_{\mathrm{dof}}}
        +
        \delta
        \mathbf{G}^{\top}
        \mathbf{G}
        &
        \mathbf{A}^{\top}
        \\
        \mathbf{A}
        &
        \mathbf{0}
    \end{bmatrix},
\end{align}
and
\begin{align}
    \mathbf{r}
    =
    \begin{bmatrix}
        \delta
        \mathbf{G}^{\top}
        \left(
        \mathbf{h}
        -
        {}^{l+1}\boldsymbol{\zeta}
        \right)
        +
        {}^{l}\boldsymbol{\lambda}
        +
        \tilde{\dot{\boldsymbol{\Theta}}}
        \\
        \mathbf{b}
    \end{bmatrix}.
    \label{eq:admm_KKT}
\end{align}

\paragraph{Step 3: Bregman update}
Finally, the primal-space multiplier is updated by mapping the constraint residual back through $\mathbf{G}^{\top}$:
\begin{align}
    {}^{l+1}\boldsymbol{\lambda}
    =
    {}^{l}\boldsymbol{\lambda}
    -
    \frac{\delta}{2}
    \mathbf{G}^{\top}
    \left(
    \mathbf{G}
    {}^{l+1}\dot{\boldsymbol{\Theta}}^{*}
    -
    \mathbf{h}
    +
    {}^{l+1}\boldsymbol{\zeta}
    \right).
    \label{eq:admm_dual}
\end{align}
Unlike a conventional constraint-space ADMM multiplier, $\boldsymbol{\lambda}$ is maintained in the primal trajectory space. The residual correction therefore enters the next primal minimization directly through the right-hand side of~\eqref{eq:admm_KKT}.

\subsubsection{Computational Properties}

Several properties make this solver well suited for GPU-accelerated bimanual planning.

\paragraph{Pre-computable factorization}
The KKT matrix $\mathbf{M}$ in~\eqref{eq:admm_primal} depends only on the constant matrices $\mathbf{G}$ and $\mathbf{A}$, together with the penalty parameter $\delta$---not on the sampled trajectory $\tilde{\dot{\boldsymbol{\Theta}}}$ or the projected variable $\dot{\boldsymbol{\Theta}}^{*}$. Therefore, a factorization of $\mathbf{M}$ can be computed once before the MPC loop begins and reused for every sample, every joint, and every MPC iteration. This is the dominant cost of the solver and can be amortized entirely.

\paragraph{Joint-wise separability}
Although the derivation above is written in stacked trajectory form, the QP remains separable across joints because the objective and all derivative constraints act independently on each joint coordinate. In implementation, we exploit this structure by decomposing the full projection into $N_{\mathrm{dof}}$ independent horizon-wise QPs with identical algebraic structure. For the bimanual setting with two UR5e arms
($N_{\mathrm{dof}}=12$), each Bregman iteration reduces to $12 \times n$ independent joint-wise projection problems, where $n$ is the sampling batch size.

\paragraph{Batch parallelism}
The updates in~\eqref{eq:admm_zeta}--\eqref{eq:admm_dual} consist entirely of matrix-vector products, element-wise operations, and solves with the pre-factorized KKT matrix. All $n$ sampled trajectories in Alg.~\ref{algo_1} are processed in parallel through batched GPU operations. The only quantities that vary across samples are the raw trajectory $\tilde{\dot{\boldsymbol{\Theta}}}$ and the associated right-hand side in~\eqref{eq:admm_KKT}; the matrices $\mathbf{M}$, $\mathbf{G}$, and $\mathbf{A}$ remain fixed.

\paragraph{Boundary value update}
The vector $\mathbf{b}$ and portions of $\mathbf{h}$ depend on the boundary values
$(\boldsymbol{\theta}_{\mathrm{init}},\,\dot{\boldsymbol{\theta}}_{\mathrm{init}})$,
which change at each MPC step as new state feedback is received. However, these quantities affect only the right-hand side vectors $\mathbf{h}$ and $\mathbf{b}$, not the KKT matrix $\mathbf{M}$. Thus, updating the boundary values requires only recomputing $\mathbf{h}$ and $\mathbf{b}$---a negligible operation compared with the precomputed factorization.

\paragraph{Convergence}
The projection problem in~\eqref{eq:compact_qp} is a convex quadratic program with affine constraints. The ADMM/Bregman-splitting iterations in~\eqref{eq:admm_zeta}--\eqref{eq:admm_dual} follow the structure used in~\cite{andrejev2025pi,ghadimi2014optimal} and converge to the global optimum of the projection problem under standard assumptions. In practice, we observe that $10$--$20$ iterations are sufficient to obtain solutions with low constraint residuals for the bimanual tasks considered in Section~\ref{sec:cost_design}.

\subsection{Bayesian Optimization}\label{subsec:Bayesian_Optimization}

\noindent The performance of the sampling-based MPC planner depends critically on the choice of cost weights, denoted collectively as the vector $\mathbf{w}$, which govern the relative importance of various safety, tracking, and coordination objectives. As we will detail in the subsequent Section~\ref{sec:cost_design}, different cooperative tasks require highly specialized cost formulations. Manually tuning the corresponding weights for each task is labor-intensive, non-systematic, and often fails to generalize across varying initial conditions. To address this, we employ \textit{Bayesian Optimization} (BO) to automatically search for optimal weight configurations, as summarized in Algorithm~\ref{algorithm_bo}. We utilize the BO implementation provided in~\cite{bayex}.

The procedure maintains a Gaussian Process (GP) surrogate model that approximates the mapping from cost weights to task performance (line~1). At each BO iteration $b$, a candidate weight vector $\mathbf{w}_b$ is sampled using the Expected Improvement (EI) acquisition function (line~3) and assigned to the planner's cost function (line~4). To promote generalization across varying task configurations, the simulation environment is randomized at the start of each episode by sampling the initial robot configuration, object pose, and target pose from predefined distributions (line~5).

The candidate weights are then evaluated by executing a full episode of up to $T$ MPC steps (lines~7--14). At each step, the MPC planner (Fig.~\ref{fig:mpc_planner}) computes and applies an action (line~8), after which the simulation state is advanced (line~9). A dense cost is accumulated at every step (line~10):
\[
\ell_t^{\text{dense}} = \ell_t^{\text{collision}} + \ell_t^{\text{coord}},
\]
where $\ell_t^{\text{collision}}$ penalizes undesired contacts that occur or are anticipated over the planning horizon, and $\ell_t^{\text{coord}}$ penalizes lack of coordination between the end-effectors and the object during the \textbf{move} phase. The episode terminates early if the goal is reached (lines~11--13), recording the actual termination step $T_{\text{end}} \leq T$.

Upon episode completion, a sparse terminal cost is computed (line~15) to encode task success:
\[
\ell^{\text{sparse}} =
\begin{cases}
\ell_{\text{fail}} + \ell_{\text{pose}}, & \text{if the goal is not reached within } T, \\
\ell_{\text{pose}}, & \text{otherwise},
\end{cases}
\]
where $\ell_{\text{pose}}$ penalizes the final position and orientation error between the object and its target configuration. For the handover task, during the \textbf{home} phase, $\ell_{\text{pose}}$ is instead defined as the joint-space deviation between the final and initial manipulator configurations. Since $\ell_{\text{pose}} \ll \ell_{\text{fail}}$, unsuccessful executions incur a significantly larger penalty than successful ones. The total episode cost combines both components (line~15):
\[
J_b(\mathbf{w}_b) = \ell^{\text{sparse}} + \sum_{t=1}^{T_{\text{end}}} \ell_t^{\text{dense}}.
\]
This scalar objective captures both task success and execution quality. The observation pair $(\mathbf{w}_b, J_b)$ is used to update the GP surrogate model (line~16), which refines the acquisition function and guides the selection of subsequent candidates. After $B$ iterations, the weight vector with the lowest observed cost is returned (line~17):
\[
\mathbf{w}^\ast = \arg\min_{\mathbf{w}_b} J_b(\mathbf{w}_b).
\]
This procedure enables systematic and data-efficient tuning of cost weights, yielding consistent improvements in performance, robustness, and generalization compared to manual tuning.

\noindent
\begin{tightalgorithm}[!t]
\caption{Bayesian Optimization for Cost Weight Tuning}
\label{algorithm_bo}
\small
\SetAlgoLined
\KwIn{Task specification, Sampling-based MPC planner (Alg.~\ref{algo_1}), BO budget $B$, max episode timesteps $T$}
\KwOut{Optimal cost weights $\mathbf{w}^\ast$}

Initialize GP surrogate model and EI acquisition function \\

\For{$b=1$ \KwTo $B$}
{
    Sample candidate weights $\mathbf{w}_b$ via EI acquisition function \\

    Assign $\mathbf{w}_b$ to planner cost function \\

    Reset environment: randomize robot configuration, object pose, and target pose \\

    $T_{\text{end}} \gets T$ \\

    \For{$t=1$ \KwTo $T$}
    {
        Run MPC planner (Fig.~\ref{fig:mpc_planner}), apply resulting action \\

        Advance simulation state \\

        Accumulate $\ell_t^{\text{dense}} = \ell_t^{\text{collision}} + \ell_t^{\text{coord}}$ \\

        \If{goal reached}
        {
            $T_{\text{end}} \gets t$ \\
            \textbf{break}
        }
    }

    Compute $\ell^{\text{sparse}}$ and episode cost $J_b(\mathbf{w}_b) = \ell^{\text{sparse}} + \sum_{t=1}^{T_{\text{end}}} \ell_t^{\text{dense}}$ \\

    Update GP surrogate with $(\mathbf{w}_b, J_b(\mathbf{w}_b))$ \\
}

\Return{$\mathbf{w}^\ast = \arg\min_{\mathbf{w}_b} J_b(\mathbf{w}_b)$}

\normalsize
\end{tightalgorithm}

\section{Cost Function Design}\label{sec:cost_design}
\begin{figure}[h]
    \centering
    \includegraphics[width=0.4\textwidth]{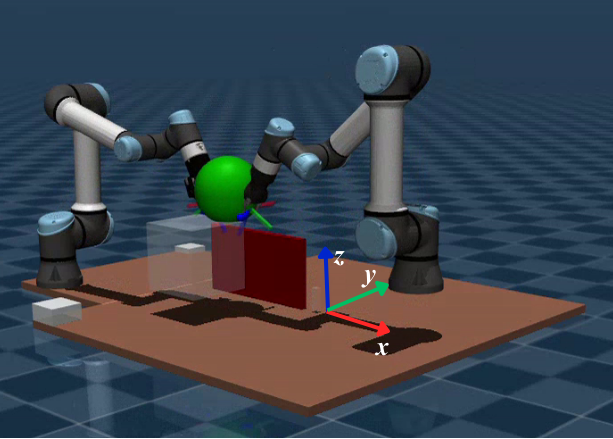}
    \caption{\footnotesize{The bimanual setup and global reference frame. The coordinate system ($x, y, z$) is explicitly modeled in the task-specific cost functions to enforce constraints, such as planar alignment in the $yz$-plane during the ball transport and handover tasks.}}
    \label{fig:robot_global_frame}
\end{figure}

This section presents one of the main contributions of our paper. We consider more challenging variants of tasks from the PerAct$^2$ benchmark and design task-specific cost functions that guide the sampling-based optimizer of Alg.~\ref{algo_1} toward successful task completion. The tasks considered in our evaluation are \textbf{Moving Tray}, \textbf{Lifting and Transporting Ball}, \textbf{Hand-Over Cubes}, and \textbf{Lifting and Transporting Box}. Each named cost component below defines a horizon-aggregated quantity $\overline{c}_i = \sum_{k=1}^{H} \overline{c}_{i,k}$, where $\overline{c}_{i,k}$ is the per-step contribution appearing in~\eqref{cost_components}. The total cost for each task is the weighted combination that enters the rollout evaluation in~\eqref{rollout_cost}.

Let the end-effector pose of the $i$-th manipulator at horizon step $k$ be denoted as $\mathbf{e}_{i,k} = (\mathbf{p}_{i,k},\, \mathbf{q}_{i,k})$, where $\mathbf{p}_{i,k} \in \mathbb{R}^3$ and $\mathbf{q}_{i,k} \in \mathbb{S}^3$ are the position and orientation (as a unit quaternion), respectively, with $i \in \{1, 2\}$. Note that $\mathbf{e}_{i,k}$ is part of the augmented environment state $\mathbf{x}_k$ obtained by rolling out the velocity sequence in the MuJoCo simulator.

\subsection{Shared Cost Components}\label{subsec:shared_costs}
\noindent Before detailing the task-specific objectives, we define three foundational cost components utilized across multiple tasks to ensure collision avoidance, joint regularization, and basic kinematic coordination.

\paragraph{Collision Cost}
Let $\mathbf{d}_k \in \mathbb{R}^{B}$ denote the concatenation of distances at horizon step $k$ between the two manipulators and $B$ contact points. Collision avoidance is ensured if $\mathbf{d}_k \geq 0,\; \forall k$. Define $\mathbf{g}_k = -\mathbf{d}_{k+1} + (1 - \gamma)\, \mathbf{d}_k$ for $\gamma \in (0, 1)$. The collision cost is:
\begin{align}
    \overline{c}_c = \sum_{k=1}^{H-1} \max(\mathbf{g}_k, 0) + \sum_{k=1}^{H} \mathbf{1}_{\{\mathbf{d}_k < 0\}},
    \label{collision_cost}
\end{align}
where $\mathbf{1}$ is an indicator function. The second term is the conventional collision cost that penalizes penetration. The first term, inspired by barrier functions~\cite{zeng2021safety}, prevents $\mathbf{d}_k$ from reaching the boundary of the infeasible region by repelling the arms from contacting bodies along the planned trajectory.

\paragraph{Joint Deviation Cost}
Let $\boldsymbol{\theta}_{\text{home}}$ denote the home joint configuration of both manipulators. We penalize deviation from this configuration:

\begin{align}
    \overline{c}_\theta = \sum_{k=1}^{H} \left\| \boldsymbol{\theta}_k - \boldsymbol{\theta}_{\text{home}} \right\|_2.
    \label{home_cost}
\end{align}

\paragraph{Relative Velocity Cost}
When both manipulators are coupled to a rigid body, they must move in a coordinated fashion. We model this requirement as:
\begin{align}
    \overline{c}_v = \sqrt{\sum_{k=1}^{H} \left( \big(\mathbf{p}_{1,k} - \mathbf{p}_{2,k}\big) \cdot \big(\dot{\mathbf{p}}_{1,k} - \dot{\mathbf{p}}_{2,k}\big) \right)^2}.
    \label{rel_vel_cost}
\end{align}
This cost forces the relative velocity between the end-effectors to be perpendicular to their relative position, preventing stretching or compression during transport.

%% ========================================================================
\subsection{Moving Tray}\label{subsec:moving_tray}
\noindent This is a more challenging version of the tray-lifting task in the PerAct$^2$ benchmark~\cite{grotz2024peract2}. The tray must be lifted and transported to a given goal pose, while avoiding collisions. We divide the task into two phases: \textbf{pick} and \textbf{move}. The former guides the manipulators toward the pick location, while the latter handles transportation to the target. 

\paragraph{End-Effector Vertical Alignment Cost}
This cost penalizes deviations in the $z$-coordinates (see Fig.~\ref{fig:robot_global_frame}) of the two end-effectors, ensuring they remain at the same height:
\begin{align}
    \overline{c}_z = \left\| \mathbf{p}_{z,1} - \mathbf{p}_{z,2} \right\|_2.
    \label{vertical_alignment_cost}
\end{align}

\paragraph{Pick Position \& Orientation Costs}
Let $\mathbf{p}_{\text{pick},i}$ and $\mathbf{q}_{\text{pick},i}$ denote the desired position and orientation (as a unit quaternion) for grasping the tray by the $i$-th arm. We define:

\small
\begin{align}
    \overline{c}_p^{\text{pick}} &= \frac{1}{2} \sum_{k=1}^{H} \Big( \|\mathbf{p}_{1,k} - \mathbf{p}_{\text{pick},1}\|_2 + \|\mathbf{p}_{2,k} - \mathbf{p}_{\text{pick},2}\|_2 \Big). \label{tray_pick_pos_cost} \\
    \overline{c}_r^{\text{pick}} &= \frac{1}{2} \sum_{k=1}^{H} \left( 2 \cos^{-1} \left| \langle \mathbf{q}_{1,k}, \mathbf{q}_{\text{pick},1} \rangle \right| + 2 \cos^{-1} \left| \langle \mathbf{q}_{2,k}, \mathbf{q}_{\text{pick},2} \rangle \right| \right). \label{tray_pick_ori_cost}
\end{align}
\normalsize

\paragraph{End-Effector Distance Cost}
Let $l_{\text{tray}}$ be the dimension of the tray along which it is held by the manipulators. The following cost ensures that both arms maintain this separation during transport:
\begin{align}
    \overline{c}_l = \sum_{k=1}^{H} \left( \|\mathbf{p}_{1,k} - \mathbf{p}_{2,k}\|_2 - l_{\text{tray}} \right)^2.
    \label{end_effector_distance_cost}
\end{align}

\paragraph{Tray Position \& Orientation Costs}
Let $\mathbf{p}_{\text{tray},k}$ and $\mathbf{q}_{\text{tray},k}$ be the position and orientation of the tray at horizon step $k$. We penalize deviation from the target pose:
\begin{align}
    \overline{c}_{p,\text{tray}} &= \sum_{k=1}^{H} \left\| \mathbf{p}_{\text{tray},k} - \mathbf{p}_{\text{target}} \right\|_2. \label{tray_position_cost} \\
    \overline{c}_{r,\text{tray}} &= \sum_{k=1}^{H} 2 \cos^{-1} \left| \langle \mathbf{q}_{\text{tray},k},\, \mathbf{q}_{\text{target}} \rangle \right|. \label{tray_ori_cost}
\end{align}

\begin{notebox}
    Note that the tray is not an actuated body. Its position and orientation can only be indirectly controlled through the manipulators.
\end{notebox}

\paragraph{Move Position \& Orientation Costs}
During the \textbf{move} phase, we define desired end-effector positions $\mathbf{p}_{\text{move},i,k}$ and orientations $\mathbf{q}_{\text{move},i,k}$ such that both end-effectors remain tightly coupled with the tray. 
\small
\begin{align}
    \overline{c}_p^{\text{move}} &= \frac{1}{2} \sum_{k=1}^{H} \Big( \|\mathbf{p}_{1,k} - \mathbf{p}_{\text{move},1,k}\|_2 + \|\mathbf{p}_{2,k} - \mathbf{p}_{\text{move},2,k}\|_2 \Big). \label{tray_move_pos_cost} \\
    \overline{c}_r^{\text{move}} &= \tfrac{1}{2} \sum_{k=1}^{H} \Big( 2 \cos^{-1}\!\big| \langle \mathbf{q}_{1,k}, \mathbf{q}_{\text{move},1,k} \rangle \big| + 2 \cos^{-1}\!\big| \langle \mathbf{q}_{2,k}, \mathbf{q}_{\text{move},2,k} \rangle \big| \Big). \label{tray_move_ori_cost}
\end{align}
\normalsize

\paragraph{Total Cost}
The total cost integrates the shared components ($\overline{c}_c, \overline{c}_\theta, \overline{c}_v$) and phase-specific weights:
\begin{align}
    \overline{c} &= w_c \, \overline{c}_c + w_\theta \, \overline{c}_\theta + w_z \, \overline{c}_z + w_v \, \overline{c}_v \notag \\
    &\quad + w^{\text{pick}} \, \big( w_p^{\text{pick}} \, \overline{c}_p^{\text{pick}} + w_r^{\text{pick}} \, \overline{c}_r^{\text{pick}} \big) \notag \\
    &\quad + w^{\text{move}} \, \big( w_l \, \overline{c}_l + w_{p,\text{tray}} \, \overline{c}_{p,\text{tray}} + w_{r,\text{tray}} \, \overline{c}_{r,\text{tray}} \notag \\
    &\quad \qquad \qquad + w_p^{\text{move}} \, \overline{c}_p^{\text{move}} + w_r^{\text{move}} \, \overline{c}_r^{\text{move}}\big).
\end{align}
The binary phase weights are set as $w^{\text{pick}} = 1,\, w^{\text{move}} = 0$ during the \textbf{pick} phase and $w^{\text{pick}} = 0,\, w^{\text{move}} = 1$ during the \textbf{move} phase. The remaining weights can be either manually tuned or optimized via Bayesian optimization, as described in Section~\ref{subsec:Bayesian_Optimization}.

%% ========================================================================
\subsection{Lifting and Transporting Ball}\label{subsec:lifting_ball}
This task is a more challenging variant of the ball-lifting task in the PerAct$^2$ benchmark~\cite{grotz2024peract2}. Both manipulators must cooperatively lift a ball by applying contact forces through their end-effectors and subsequently transport it to a specified location while avoiding obstacles. We divide this into two phases: \textbf{pick} and \textbf{move}. 

\paragraph{Phase-Split Collision Cost}
The collision cost follows the shared formulation~\eqref{collision_cost} but is computed separately for the \textbf{pick} and \textbf{move} phases as $\overline{c}_c^{\text{pick}}$ and $\overline{c}_c^{\text{move}}$. During the \textbf{move} phase, collisions between the ball and the manipulators are excluded from the cost so the end-effectors can press against its surface.

\paragraph{End-Effector Planar Alignment Cost}
We enforce that both manipulators apply contact forces at the same location in the $yz$-plane (see Fig.~\ref{fig:robot_global_frame}). Let $\mathbf{P}_{i,yz} \in \mathbb{R}^{H \times 2}$ denote the stacked $yz$-components of the end-effector positions for manipulator $i$. The alignment cost is:
\begin{align}
    \overline{c}_{yz} = \left\| \mathbf{P}_{1,yz} - \mathbf{P}_{2,yz} \right\|_F.
    \label{planar_alignment_cost}
\end{align}

\paragraph{End-Effector Orientation \& Distance Costs}
Target orientations and maintaining a fixed offset $l_{\text{ball}}$ follow the identical structural formulations defined in~\eqref{tray_pick_ori_cost},~\eqref{tray_move_ori_cost}, and~\eqref{end_effector_distance_cost}.

\paragraph{End-Effector to Object Alignment Cost}
Define the midpoint of the two end-effectors as $\mathbf{c}_k = \tfrac{1}{2}\big(\mathbf{p}_{1,k} + \mathbf{p}_{2,k}\big)$. During the \textbf{pick} phase, we penalize deviation from the ball position $\mathbf{p}_{\text{ball}}$ (with a small vertical offset $\epsilon$):
\begin{align}
    \overline{c}_{\text{eef-obj}} = \sum_{k=1}^{H} \left\| \{\mathbf{c}_k + (0,0,\epsilon)\} - \mathbf{p}_{\text{ball}} \right\|_2.
    \label{ball_alignment_cost}
\end{align}

\paragraph{Object-to-Target Cost}
Let $\mathbf{p}_{\text{target}}$ be the desired ball position. The transportation cost is:
\begin{align}
    \overline{c}_{\text{obj-targ}} = \sum_{k=1}^{H} \left\| \{\mathbf{c}_k + (0,0,\epsilon)\} - \mathbf{p}_{\text{target}} \right\|_2.
\end{align}

\paragraph{Total Cost}
\begin{align}
    \overline{c} =& \; w^{\text{pick}} \big(w_c \, \overline{c}_c^{\text{pick}} + w_{\text{eef-obj}} \, \overline{c}_{\text{eef-obj}}\big) \notag \\
    &+ w^{\text{move}} \big(w_c \, \overline{c}_c^{\text{move}} + w_{\text{obj-targ}} \, \overline{c}_{\text{obj-targ}}\big) \notag \\
    &+ w_\theta \, \overline{c}_\theta + w_{yz} \, \overline{c}_{yz} + w_v \, \overline{c}_v + w_r \, \overline{c}_r + w_l \, \overline{c}_l.
\end{align}

%% ========================================================================
\subsection{Hand-Over Task}\label{subsec:handover}
\noindent This task is taken from the PerAct$^2$ benchmark~\cite{grotz2024peract2} but made more challenging by the addition of static obstacles. We divide it into four phases: \textbf{pick}, \textbf{pass}, \textbf{place}, and \textbf{home}. In the \textbf{pick} phase, the assigned manipulator approaches and grasps the object. The \textbf{pass} phase facilitates a coordinated handover, transferring the object between manipulators under geometric and alignment constraints. During the \textbf{place} phase, the receiving manipulator transports the object to the desired target location. Finally, in the \textbf{home} phase, both manipulators return to their initial joint configurations.

\paragraph{End-Effector Planar Alignment Cost}
During the \textbf{pass} phase, we require both manipulators to be aligned in the $yz$-plane using the shared alignment cost defined in~\eqref{planar_alignment_cost}.

\paragraph{Phase-Specific Position Costs}
Let $\mathbf{p}_{\text{obj}}$, $\mathbf{p}_{\text{pass},i}$, and $\mathbf{p}_{\text{target}}$ denote the desired positions for the \textbf{pick}, \textbf{pass}, and \textbf{place} phases of the $i$-th manipulator. 
\begin{align}
    \text{Pick:} \quad & \overline{c}_{p,i}^{\text{pick}} = \sum_{k=1}^{H} \big\| \mathbf{p}_{i,k} - (\mathbf{p}_{\text{obj}} + [0,0,\epsilon]) \big\|_2, \\[4pt]
    \text{Pass:} \quad & \overline{c}_{p,i}^{\text{pass}} = \sum_{k=1}^{H} \sum_{d \in \mathcal{D}} \big| \mathbf{p}_{(d)i,k} - \mathbf{p}_{(d)\text{pass},i} \big|, \\[4pt]
    \text{Place:} \quad & \overline{c}_{p,i}^{\text{place}} = \sum_{k=1}^{H} \big\| \mathbf{p}_{i,k} - \mathbf{p}_{\text{target}} \big\|_2,
\end{align}
where $\mathcal{D} \subseteq \{x, z\}$ denotes the set of active dimensions that define the handover position. The combined position cost is $\overline{c}_p = \sum_{i=1}^{2} \Big( w_{i}^{\text{pick}} \, \overline{c}_{p,i}^{\text{pick}} + w_{i}^{\text{pass}} \, \overline{c}_{p,i}^{\text{pass}} + w_{i}^{\text{place}} \, \overline{c}_{p,i}^{\text{place}} \Big)$.

\paragraph{Phase-Specific Orientation Costs}
For each manipulator and phase, we define a target unit quaternion $\mathbf{q}_{\text{target},i}^{\text{phase}}$. The combined orientation cost is:
\begin{align}
    \overline{c}_{r,i}^{\text{phase}} &= \sum_{k=1}^{H} 2 \cos^{-1} \Big( \big| \langle \mathbf{q}_{i,k},\, \mathbf{q}_{\text{target},i}^{\text{phase}} \rangle \big| \Big). \\
    \overline{c}_r &= \sum_{i=1}^{2} \Big( w_{i}^{\text{pick}} \, \overline{c}_{r,i}^{\text{pick}} + w_{i}^{\text{pass}} \, \overline{c}_{r,i}^{\text{pass}} + w_{i}^{\text{place}} \, \overline{c}_{r,i}^{\text{place}} \Big).
\end{align}

\paragraph{Home Return Cost}
During the \textbf{home} phase, an additional cost drives each manipulator back to its initial joint configuration:
\begin{align}
    \overline{c}_{\text{home},i} = \sum_{k=1}^{H} \big\| \boldsymbol{\theta}_{i,k} - \boldsymbol{\theta}_{\text{home},i} \big\|_2,
\end{align}

\paragraph{Total Cost}
The overall cost is:
\begin{align}
    \overline{c} = &\; w_c \, \overline{c}_c + w_{\theta} \, \overline{c}_\theta + w^{\text{pass}} w_{yz} \, \overline{c}_{yz} \nonumber \\ 
    &+ w_p \, \overline{c}_p + w_r \, \overline{c}_r + \sum_{i=1}^{2} w_{i}^{\text{home}} \, \overline{c}_{\text{home},i}.
\end{align}

\subsection{Lifting and Transporting Box}\label{subsec:lifting_box}
\noindent This task is a geometric variant of the ball-lifting task described in Section~\ref{subsec:lifting_ball}, where the spherical payload is replaced with a rigid box. The inherent asymmetry of the box makes this task significantly more challenging, as the object is susceptible to rotational slipping and misalignment. Both manipulators must apply continuous, precisely coordinated contact forces to maintain a stable hold. 

\paragraph{Box-Specific Alignment}
The cost components extend the formulation established for the ball, heavily modified for flat-sided geometry. First, the box requires individual alignment of each end-effector to designated contact locations $\mathbf{p}_{\text{box},i}$ on the object's opposing faces. 
\begin{align}
    \overline{c}_{\text{eef-obj}} = \sum_{k=1}^{H} \frac{1}{2} \Big( \|\mathbf{p}_{1,k} - \mathbf{p}_{\text{box},1}\|_2 + \|\mathbf{p}_{2,k} - \mathbf{p}_{\text{box},2}\|_2 \Big).
\end{align}

\paragraph{Box orientation}
Second, an additional orientation cost is introduced during the \textbf{move} phase to regulate the box orientation and prevent undesired rotations of the non-symmetric payload. Due to the increased complexity of this task, we do not introduce additional obstacles in this setting.

\begin{notebox}
\noindent\textbf{Runtime Phase Transitions:} {Phase transitions during execution are determined dynamically at runtime rather than relying on fixed time steps. At each control step, the system continuously evaluates the state-dependent cost functions, specifically the positional distances  and rotational quaternion distances between the robotic end-effectors and the target objects. A phase transition (for example, switching from a \textit{pick} phase to a \textit{move} phase) is triggered strictly when all relevant positional and rotational costs fall below predefined acceptable thresholds.
This ensures that the robot only progresses to the next stage of the task once a stable, successful grasp or manipulation is confirmed by the system state.}
\end{notebox}

\section{Validation and Benchmarking}\label{subsec:metric}

The objective of this section is two-fold:
\begin{itemize}
    \item To validate every component of our pipeline: sampling-based optimizer, MPC and Bayesian cost tuning.
    \item To demonstrate how our approach improves upon existing sampling-based optimizers in bimanual tasks.
\end{itemize}

\subsection{Implementation Details}
\begin{figure}[!t]
    \centering
    \includegraphics[width=0.85\linewidth]{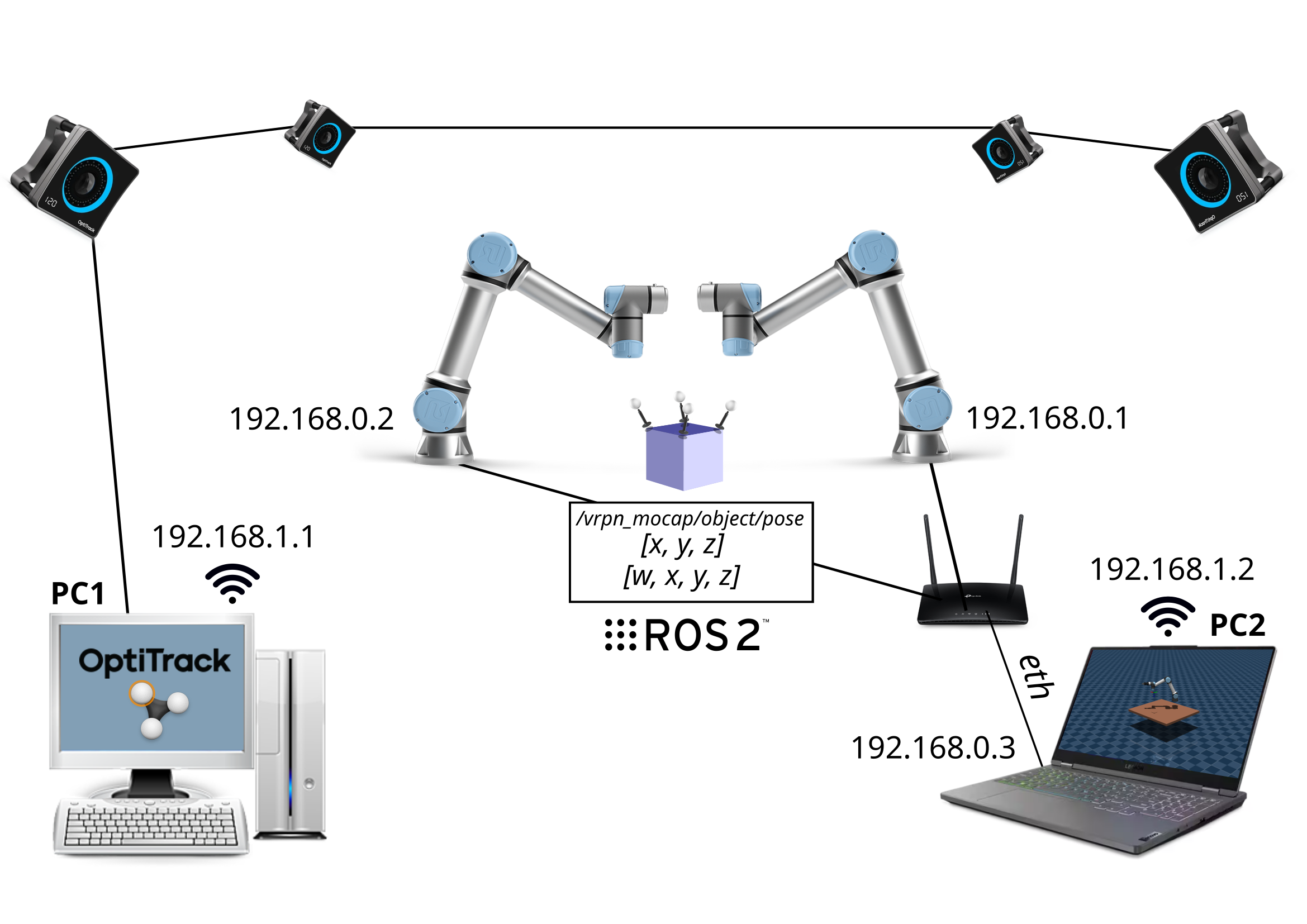}
    \caption{\footnotesize{Experimental Setup wherein real-world and simulation are tightly coupled with OptiTrack motion capture.}}
    \label{fig:real_life_setup}
    \vspace{-2em}
\end{figure}
\noindent Alg.~\ref{algo_1} was implemented in JAX with the GPU-accelerated MJX physics engine for receding horizon control. The simulation, featuring two UR5e manipulators with Robotiq grippers, uses a 0.1s time-step that was experimentally chosen to balance accuracy and planning time. High-fidelity sim-to-real transfer was achieved by meticulously aligning the physical and simulated environments and continuously updating the simulation to reflect the real world. The implementation involved the following hardware and data pipeline (Fig.~\ref{fig:real_life_setup}). A unified network was established by connecting both manipulators and a control laptop (PC2) to a single Wi-Fi router using Ethernet, ensuring reliable, low-latency command and feedback. The control laptop was responsible for running the MuJoCo simulation and our proposed planner. In addition, the control laptop was wirelessly connected to the same Wi-Fi network as the PC1 that controlled the OptiTrack system, which broadcast the reception of the pose data of dynamic objects (obstacles, ball, tray, etc.) through ROS2 topics. The control laptop subscribed to these topics, integrating the incoming OptiTrack data directly into the simulation by updating the Mocap body features within MuJoCo. This mechanism synchronized the simulation with the physical workspace in real-time.

\vspace{-1em}

\subsection{Qualitative Results}
\noindent In this subsection, we describe the simulation and hardware runs achieved with our approach. 

Figure~\ref{fig:moving_tray_snapshots} illustrates a time-lapse sequence of the Moving Tray task executed in both the physics simulation and on the physical hardware. The dual-arm system successfully transitions from the \textbf{pick} phase to the \textbf{move} phase. During transport, the manipulators maintain a stable, rigid grasp on the tray, keeping it level without dropping it. This demonstrates the effectiveness of the aforementioned cost components in successfully guiding the arms and the tray safely around and between the wall-like obstacles to reach the final target pose.

\begin{figure*}[h]
    \centering
    \includegraphics[width=\linewidth]{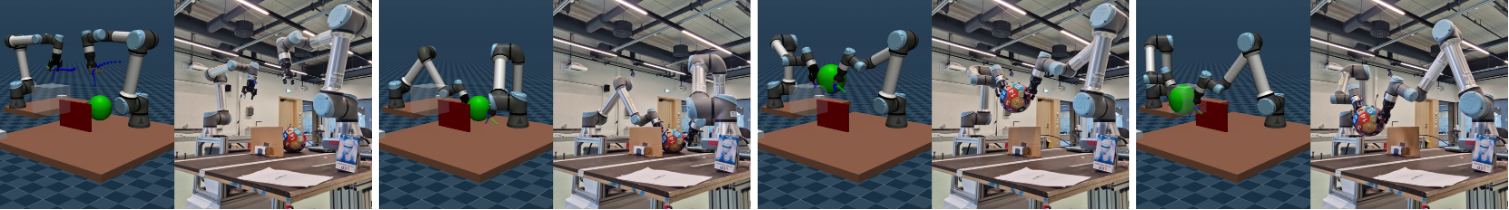}
    \caption{\footnotesize{ Execution of the ball lifting and transport task. The manipulators must apply continuous, opposing contact forces in the $yz$-plane to maintain the unattached ball while dynamically adjusting their joint configurations to clear the central obstacle.}}
    \label{fig:moving_ball_snapshots}
\end{figure*}

Figure~\ref{fig:moving_ball_snapshots} captures the execution of the ball-lifting task across both simulation and real-world environments. This task demands continuous, carefully controlled contact during the \textbf{move} phase. The snapshots confirm that the manipulators dynamically adjust to maintain this hold while navigating a complex workspace, effectively clearing the central obstacle structures. 

\begin{figure*}[h]
    \centering
    \includegraphics[width=\linewidth]{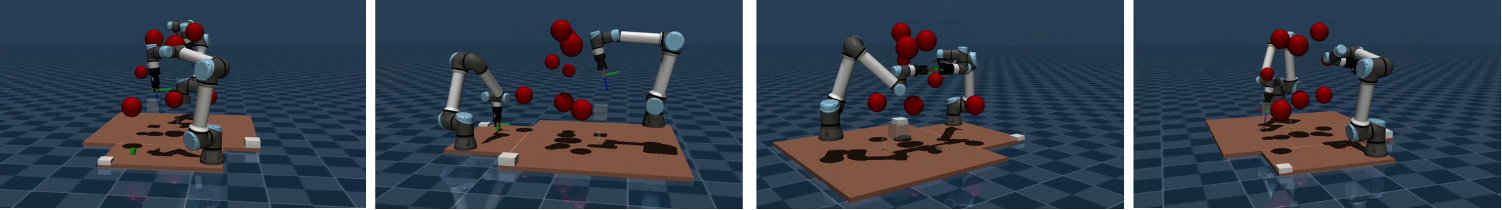}
    \caption{\footnotesize{Execution of the handover task. The sequence demonstrates the transition from the pick phase to the pass phase, requiring precise mid-air synchronization and collision avoidance amidst a dense cluster of static spherical obstacles.}}
    \label{fig:handover_snapshots}
\end{figure*}

The sequential execution of the handover task is demonstrated in Figure~\ref{fig:handover_snapshots}. Unlike continuous transport, this maneuver relies on independent but highly synchronized manipulator roles. The sequence illustrates the first arm successfully navigating through a dense field of suspended obstacles to retrieve the green cube. The planner seamlessly transitions from \textbf{pick} to \textbf{pass} phase, driving both end-effectors to a precise mid-air point for the exchange, then proceeds to the \textbf{place} and \textbf{home} phases.

\begin{figure*}[h]
    \centering
    \includegraphics[width=\linewidth]{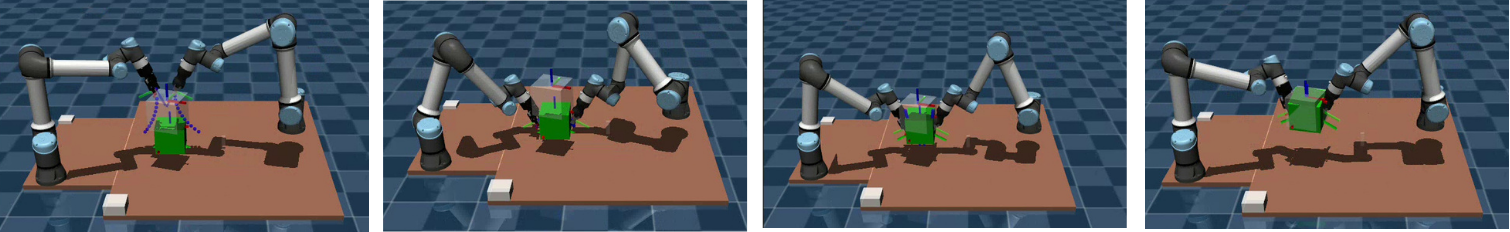}
    \caption{\footnotesize{Execution of the box lifting task. Unlike the spherical payload, the asymmetric flat-sided box requires the optimizer to enforce strict orientation costs during the move phase to prevent rotational slipping while sustaining frictional contact.}}
    \label{fig:lift_box_snapshots}
\end{figure*}

Figure~\ref{fig:lift_box_snapshots} demonstrates the simulated execution of the box-lifting task. Guided by the cost components, the manipulators successfully target the opposing flat faces of the green box during the \textbf{pick} phase to establish a secure frictional grasp. The visual sequence confirms that the newly introduced orientation costs effectively prevent any unwanted rotation or tilting during \textbf{move} phase.

\subsection{Statistical Validation}\label{sec:quantitative_results}
\begin{figure*}
    \centering
    \includegraphics[width=\textwidth]{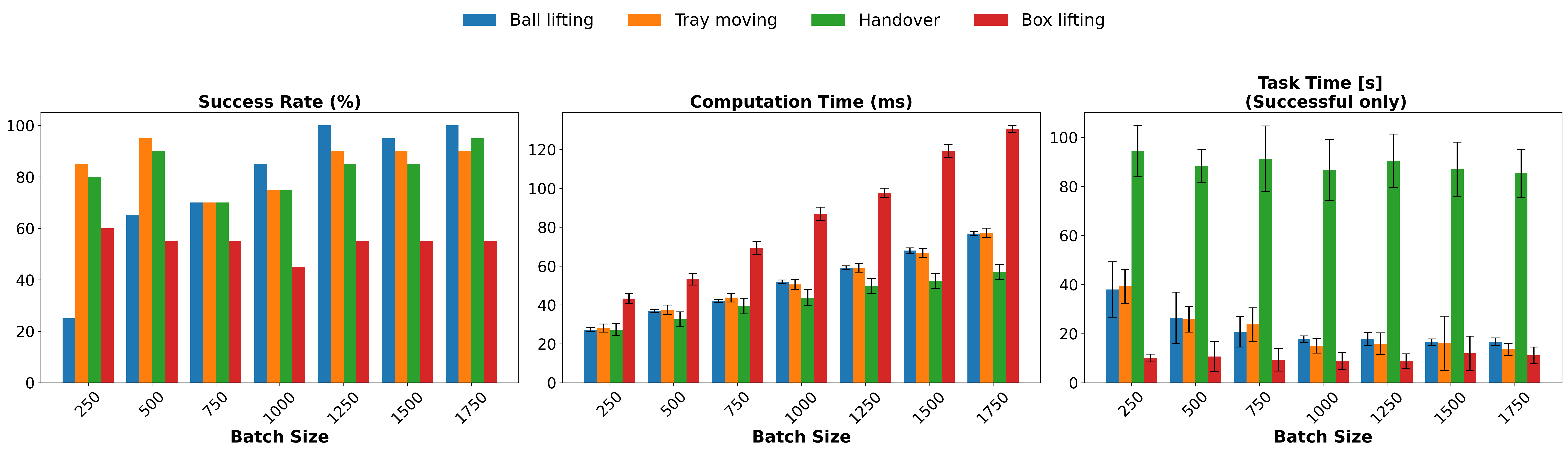}
    \caption{\footnotesize{Statistics of all tasks. (a) Success rate, (b) computation time, and (c) task time with respect to different batch sizes used in Alg.\ref{algo_1}. Note that success rates generally plateau after a batch size of 1250, while the computation time for the asymmetric box lifting task scales significantly higher than other tasks. }}
    \label{fig:stat_all_task}
\end{figure*}

\noindent In this subsection, we present the statistical validation of our approach. To this end, we run Alg.\ref{algo_1} by randomizing the goal pose in each task. The evaluation is conducted over varying batch sizes (i.e., number of samples in Alg.~\ref{algo_1}), with statistics computed over 20 runs per setting (Fig.~\ref{fig:stat_all_task}). We ground our analysis in the following metrics.

\noindent \textbf{Success Metric}
\begin{align}
\text{Success Rate (\%)} = \frac{N_{\mathrm{success}}}{N_{\mathrm{total}}} \times 100
\end{align}
where $N_{\mathrm{success}}$ is the number of successful runs and $N_{\mathrm{total}}$ is the total number of experiments of a task. Failure occurs when there is significant collision cost in the planning horizon, the task is not completed within pre-defined threshold time, or the object falls during `move' phase.\\

\noindent \textbf{Mean and Standard Deviation of Task Time}
\begin{equation}
\begin{aligned}
\bar{t}_{\mathrm{task}} &= \frac{1}{N_{\mathrm{runs}}} \sum_{i=1}^{N_{\mathrm{runs}}} t_{\mathrm{task}}^{(i)}, \\
\sigma_{\mathrm{task}} &= \sqrt{\frac{1}{N_{\mathrm{runs}}} \sum_{i=1}^{N_{\mathrm{runs}}} 
\left( t_{\mathrm{task}}^{(i)} - \bar{t}_{\mathrm{task}} \right)^2 }
\end{aligned}
\end{equation}
where $t_{\mathrm{task}}^{(i)}$ is the total time to complete the task in run $i$ and $N_{\mathrm{runs}} = 20$.

\noindent \textbf{Mean and Standard Deviation of Computation Time}
% Computation time
\begin{equation}
\begin{aligned}
\bar{t}_{\mathrm{comp}} &= \frac{1}{N_{\mathrm{steps}}} \sum_{m=1}^{N_{\mathrm{steps}}} t_{\mathrm{comp}}^{(m)}, \\
\sigma_{\mathrm{comp}} &= \sqrt{\frac{1}{N_{\mathrm{steps}}} \sum_{m=1}^{N_{\mathrm{steps}}} 
\left( t_{\mathrm{comp}}^{(m)} - \bar{t}_{\mathrm{comp}} \right)^2 }
\end{aligned}
\end{equation}
where $t_{\mathrm{comp}}^{(m)}$ is the computation time for the $m^{th}$ planning step and $N_{\mathrm{steps}}$ is the total number of re-planning needed for the task completion.\\

The results are summarized in Fig.\ref{fig:stat_all_task}. Ball lifting shows a strong dependence on batch size, improving from $65\%$ at 500 to nearly $100\%$ for batch sizes $\geq 1250$. Tray moving maintains consistently high performance ($\sim85$–$95\%$) across all batch sizes. The hand-over task exhibits similar robustness, reaching up to $95\%$ success at larger batch sizes. In contrast, box lifting remains comparatively challenging, with success rates saturating around $55$–$60\%$, highlighting the increased difficulty due to asymmetric object geometry.

Computation time increases with batch size for all tasks. Ball lifting, tray moving, and hand-over tasks remain below $\sim80$ ms even at the largest batch sizes, enabling real-time execution. Box lifting incurs a higher computational cost (up to 130 ms) because its increased task complexity necessitates more optimization iterations to reliably obtain feasible trajectories, resulting in higher overall computation time

The reported task completion times correspond only to successful episodes. Task completion time generally decreases with increasing batch size for ball lifting and tray moving, indicating improved solution quality with more samples. The hand-over task shows relatively stable execution times ($\sim85$–$95$ s), suggesting that coordination dominates performance over sampling density. Box lifting achieves the lowest execution times ($\sim8$–$15$ s) among successful episodes, indicating that successful executions are efficient. However, the comparatively lower success rate highlights the difficulty in consistently achieving such successful outcomes.

\begin{notebox}
    Summary : Larger batch sizes generally improve success rates and task completion efficiency while maintaining real-time computation speeds for most tasks. However, manipulating asymmetric objects like boxes remains a notable challenge, yielding lower success rates and higher computational costs despite fast execution times when successful.
\end{notebox}

\subsection{Effect of Bayesian Optimization}

% \textbf{BO Tuning Configuration and Computational Cost:} 

{Bayesian Optimization (BO) was conducted over 1,000 episodes per task. To ensure that the obtained weights generalizes to diverse task configurations, we randomized robot joint configurations, object poses, and target locations at the start of each episode. On our setup (RTX 4090), the total wall-clock time for a full 1,000-episode BO run is between 3 to 9 hours, depending significantly on task complexity and the required episode length. Specifically, the duration is proportional to the number of MPC timesteps defined for each task; for instance, the \textit{Tray Move} task requires only 90 steps per episode, resulting in faster optimization, whereas the \textit{Handover} task requires up to 500 steps to accommodate the sequence of sub-tasks, leading to longer tuning times. This offline tuning cost is a one-time overhead that yields significant online reliability}.

\begin{figure*}
    \centering
    \includegraphics[width=\textwidth]{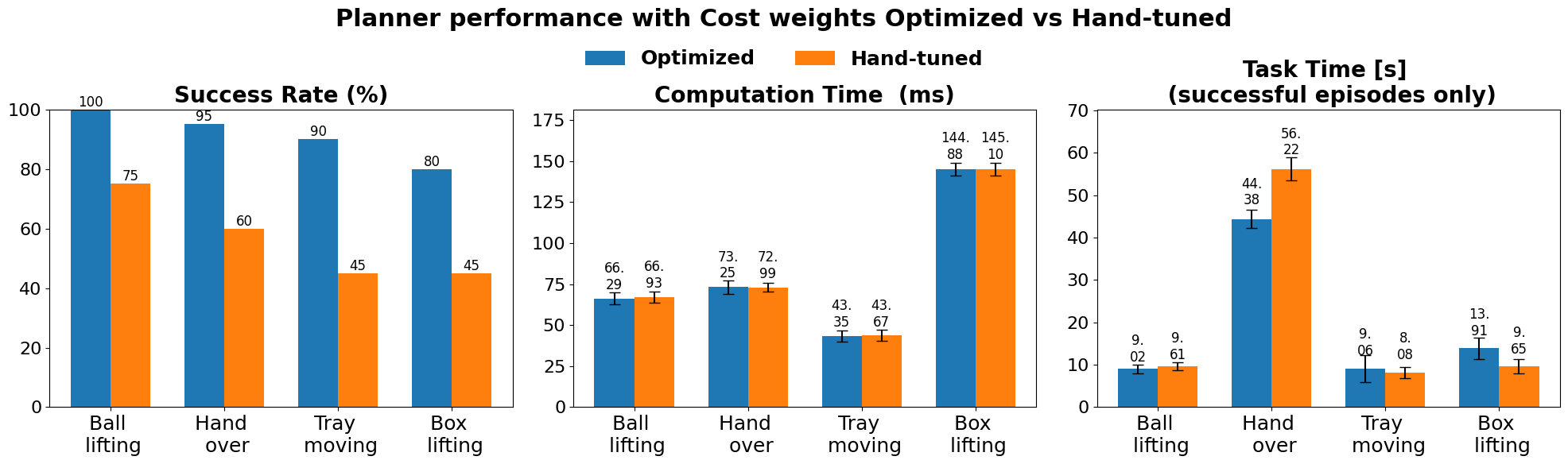}
     \caption{\footnotesize{Comparison across all tasks between manually tuned and 
    Bayesian-optimized cost weights for (a) Success rate, (b) computation time, and (c) task completion time. Bayesian optimization systematically improves success rates across all tasks—most notably a $45\%$ increase in the tray moving task—without introducing additional computational overhead during online execution.}}
    \label{fig:cw_stat_combined}
\end{figure*}

\noindent Figure~\ref{fig:cw_stat_combined} presents a quantitative comparison between manually tuned and Bayesian Optimization (BO)-optimized cost weights across multiple bimanual manipulation tasks. The evaluation considers three key metrics: success rate, computation time, and task completion time.

The BO-optimized weights improve success rates across all four tasks. Specifically, success increases from $75\%$ to $100\%$ for ball lifting, from $60\%$ to $95\%$ for handover, from $45\%$ to $90\%$ for tray moving, and from $45\%$ to $80\%$ for box lifting. These results indicate that the planner is sensitive to the relative weighting of task, coordination, and safety objectives, and that Bayesian optimization provides an effective offline mechanism for identifying more reliable cost configurations across randomized task instances.

Importantly, these improvements are achieved without introducing additional computational overhead. The average computation time remains comparable between the two approaches across all tasks, demonstrating that the optimization of cost weights does not compromise real-time feasibility. The difference in computation time across tasks is due to different number of cem iterations and planning horizon length.

Furthermore, the task completion time for successful episodes is either reduced or remains similar when using optimized weights. Notably, in the handover task, the completion time decreases significantly, indicating more efficient execution under BO-tuned parameters. For other tasks, the differences are marginal, suggesting that improved success rates are not achieved at the expense of significantly slower execution. It is important to note that task completion time was not explicitly included as an optimization objective during Bayesian Optimization; therefore, no consistent improvement in execution time is expected. The observed variations are a byproduct of improved task feasibility rather than direct optimization for speed.

\begin{notebox}
    Summary: Overall, these results demonstrate that Bayesian Optimization provides a systematic, data-efficient approach to tuning cost weights, achieving significantly higher success rates while maintaining comparable computational efficiency and execution speed. This highlights the practical advantage of BO in enhancing the robustness and generalization of sampling-based planners.

\end{notebox}

\subsection{Comparison with Existing Approaches}\label{subsec:Baseline comparison}
 
\noindent In this section, we evaluate our sampling-based optimizer with QP-based projection against three representative alternatives. We first summarize the comparison methods, then present the task-level results and connect the observed performance gaps to the structural analysis of Section~\ref{subsec:Role of QP}.
 
\subsubsection{Comparison Methods}
We consider the following approaches, each representing a distinct strategy for handling the smoothness--exploration trade-off:
 
\noindent\textit{Baseline-CEM~\cite{baseline_cem}:}
The standard CEM formulation in which raw Gaussian samples are used directly as joint velocity sequences without any smoothing or projection. This serves as the lower bound on smoothness: trajectories are uncorrelated across time steps, and their finite-difference derivatives exhibit rapidly increasing variance with derivative order (cf.\ Section~\ref{subsec:Role of QP}).
 
\noindent\textit{Low-Pass Filter (LPF)~\cite{kicki2025low}:}
A second-order Butterworth filter is applied to the sampled perturbations before rollout, attenuating high-frequency components. The cutoff frequency is a tunable hyperparameter that controls the smoothness--responsiveness trade-off. As established in Section~\ref{subsec:Role of QP}, this method cannot enforce strict derivative bounds or boundary conditions.
 
\noindent\textit{Savitzky--Golay Filter (SGF)~\cite{williams2018information}:}
A polynomial smoothing filter is applied post-hoc to the sampled control sequences. The polynomial order governs the degree of smoothing. Like LPF, SGF cannot guarantee satisfaction of kinematic bounds and does not enforce boundary continuity across MPC iterations.
 
\noindent\textit{QP Projection (ours):}
The projection defined in \eqref{proj_cost}--\eqref{proj_boundary}, which minimally corrects each sample to satisfy strict bounds on position, velocity, acceleration, and jerk while enforcing boundary conditions. As shown in Section~\ref{subsec:Role of QP}, this permits effective covariances more than 1.5 orders of magnitude larger than the alternatives.
 
All methods are evaluated on the four bimanual tasks described in Section~\ref{sec:cost_design}---ball lifting, tray moving, hand-over, and box lifting---using the three metrics defined in Section~\ref{sec:quantitative_results}: success rate, task completion time (for successful episodes), and computation time per MPC step. Statistics are collected over 20 runs per task with randomized goal poses.
 
\subsubsection{Practical Considerations}
 
An important observation is that baseline-CEM, LPF, and SGF cannot be deployed with the same minimal setup as the QP-based approach. Because these methods do not enforce strict kinematic feasibility, their sampled trajectories often exhibit jerky and dynamically inconsistent behavior, particularly under the high covariance settings needed for adequate exploration. In practice, we found that two additional modifications were necessary for these methods to reliably complete the tasks in Section~\ref{sec:cost_design}: (i) collision-free inverse kinematics~\cite{Zakka_Mink_Python_inverse_2026} with pre-pick offsets to stabilize the manipulators prior to interaction, and (ii) an explicit smoothness cost penalizing joint accelerations, appended to the task-specific cost functions. Without these modifications, the manipulators frequently failed to converge to stable grasping configurations. The QP-based approach requires neither of these additions, as kinematic feasibility and boundary continuity are guaranteed by construction.
 
Furthermore, LPF and SGF each introduce an additional hyperparameter---the cutoff frequency and polynomial order, respectively---that critically influences task performance. Lower values produce smoother trajectories but attenuate velocities, leading to sluggish motion. Higher values preserve responsiveness but reintroduce oscillations. The optimal setting is task-dependent and lacks a principled selection procedure, unlike our QP formulation where the user directly specifies physically meaningful actuator bounds.

\begin{notebox}
   We conducted an extensive search to determine the maximum control sampling covariance for the baseline-CEM, SGF, and LPF methods that still resulted in robust task execution. To ensure a fair comparison, we did not enforce strict derivative bounds on these baselines, as their original formulations were not designed to support them
\end{notebox}
 
\subsubsection{Results and Analysis}
 
\begin{figure*}
     \centering
    \includegraphics[width=\textwidth]{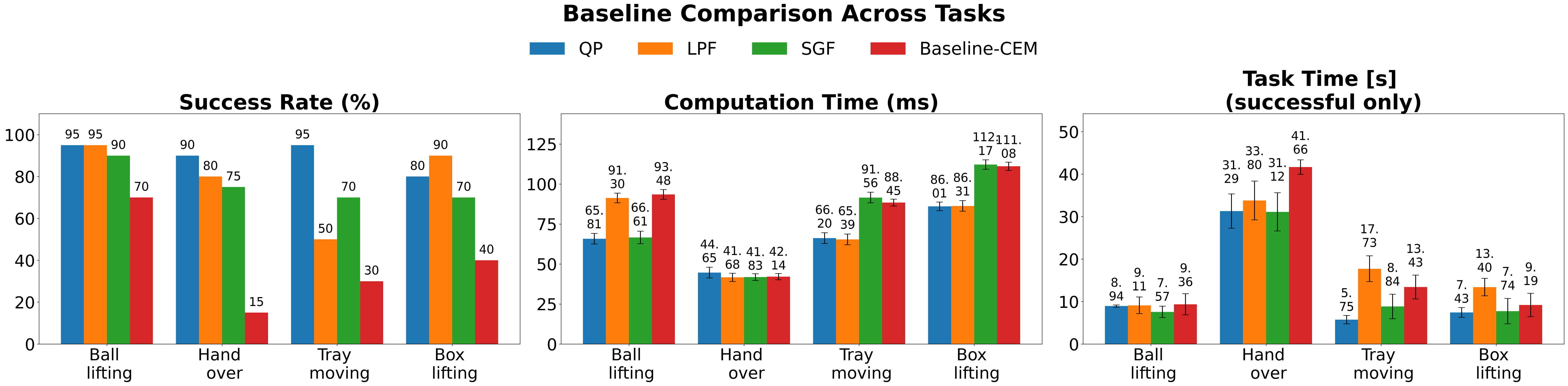}
     \caption{\footnotesize{Comparison across all tasks with existing approaches for (a) Success rate, (b) computation time, and (c) task completion time. Baseline-CEM refers to the standard unfiltered formulation of CEM~\cite{baseline_cem}. The proposed QP-projection method consistently outperforms vanilla CEM, LPF, and SGF baselines in success rate, particularly in multi-phase tasks requiring sustained bimanual coordination}}
    \label{fig:baseline_comp_stat}
    \vspace{-0.6cm}
\end{figure*}

The quantitative comparison is presented in Fig.~\ref{fig:baseline_comp_stat}. We organize the analysis by metric.
 
\paragraph{Success Rate}
The QP-based method achieves the highest or joint-highest success rate in three of the four tasks. For ball lifting, QP and LPF both attain $95\%$, followed by SGF ($90\%$) and vanilla CEM ($70\%$). In the handover task, QP reaches $90\%$, outperforming LPF ($80\%$), SGF ($75\%$), and significantly surpassing vanilla CEM ($15\%$). The most pronounced improvement appears in tray moving, where QP achieves $95\%$ compared to SGF ($70\%$), LPF ($50\%$), and baseline-CEM ($30\%$). The poor performance of baseline-CEM in the handover and tray tasks, which require precise, sustained coordination between both arms, is a direct consequence of the exploration--smoothness coupling identified in Section~\ref{subsubsec:decoupling}: the unfiltered sampler cannot explore broadly enough to discover coordinated trajectories within the available sample budget, while increasing the covariance leads to oscillatory rollouts that destabilize the MPC warm-start.
 
In the box lifting task, LPF slightly outperforms QP ($90\%$ vs.\ $80\%$), with both methods maintaining a clear advantage over SGF ($70\%$) and baseline-CEM ($40\%$). We attribute LPF's advantage here to the fact that the box task does not include obstacles, reducing the need for the aggressive exploration that the QP uniquely enables. In this simpler geometric setting, the frequency-domain smoothing of LPF provides sufficient trajectory quality.

\paragraph{Computation Time}
The QP-based method incurs computation times comparable to all alternatives. The per-iteration overhead of the QP projection is negligible (due to the use of a specialized solver described in Section \ref{subsec:custom_qp_solver}) relative to the cost of MuJoCo rollouts, which dominate the computational budget. The observed differences across methods are primarily attributable to variations in the number of optimization iterations required for convergence and the planning horizon length, rather than to the smoothing mechanism itself. This confirms that the performance gains of the QP projection are achieved without sacrificing real-time feasibility.

\paragraph{Task Completion Time}
Task completion times or task times are reported only for successful episodes. The QP-based method achieves faster or comparable execution across all tasks. This is a natural consequence of the decoupling property: because the QP projects onto the full feasible envelope (Section~\ref{subsec:Role of QP}), the resulting trajectories can exploit the maximum permissible velocities and accelerations, rather than being artificially attenuated by conservative covariance settings or frequency-domain filtering.

 The box lifting task highlights a clear trade-off: LPF achieves a higher success rate, but with substantially longer task completion times.
%A notable exception is the box lifting task, where LPF's higher success rate comes at the cost of substantially longer task completion times. 
This trade-off is characteristic of filtering-based approaches: the attenuation required to suppress derivative violations also reduces the peak velocities achievable by the manipulators, leading to slower execution. The QP avoids this trade-off entirely—it bounds derivatives without attenuating the signal within the feasible region.
 
\subsubsection{Connection to Section~\ref{subsec:Role of QP}}
 
The task-level results corroborate the qualitative predictions
of the effective-covariance analysis in
Section~\ref{subsec:Role of QP}. While the exact numerical
values of $\sigma_{\mathrm{eff}}$ reported there correspond to
a single-joint test setting, the underlying mechanism transfers
directly: baseline-CEM and LPF must operate with highly
conservative covariances to avoid highly jerky motions, sampling
almost exclusively near the current distribution mean. We reiterate that non-smooth controls destabilize the sampling-based optimization and MPC itself. In tasks
requiring precise bimanual coordination through cluttered
environments---where feasible trajectories occupy small, isolated
regions of the high-dimensional joint-velocity space---this
restricted coverage translates directly into lower success rates.
The QP projection, by contrast, permits orders-of-magnitude
larger effective covariances by construction, spanning a
substantially larger portion of the kinematically reachable set
at every planning step. This enables the optimizer to discover
high-quality coordinated trajectories within a fixed sample
budget. Furthermore, the boundary condition enforcement
in~\eqref{proj_boundary} stabilizes the MPC warm-start across
iterations, preventing the inter-step oscillations that degrade
vanilla CEM's performance in multi-phase tasks such as handover
and tray transport.
 
\subsection{Comparison with Learning-Based Approaches}

\noindent Our framework is not intended as a direct replacement for end-to-end learning methods such as~\cite{gkanatsios20253d}; rather, the two paradigms operate under different assumptions regarding data availability, task specification, and online computation. A rigorous quantitative comparison is further complicated by differences in simulator backends, manipulator models, and training distributions. We therefore present a preliminary qualitative study whose purpose is narrower: to illustrate a setting in which online model-based planning can adapt to a geometric change that lies outside the conditions used to train a learned policy.

We consider a ball-lifting task similar to that studied in~\cite{gkanatsios20253d}, but introduce a static obstacle that requires a modified lifting strategy. The scenario is recreated in MuJoCo using a Franka-Panda manipulator to obtain a consistent evaluation environment for our planner and the comparison setup. In this obstacle-shift setting, the learned policy partially adapts during the approach phase by moving around the obstacle, but often fails during the lift itself, where the required avoidance maneuver differs from the demonstrated training distribution. In contrast, PRISM evaluates candidate motions against the current obstacle configuration through online physics rollouts and can discover a feasible lifting strategy without retraining.

Over 20 trials with slightly perturbed obstacle placements, PRISM succeeds in $60\%$ of the cases, compared with $20\%$ for~\cite{gkanatsios20253d}. These numbers should not be interpreted as a definitive benchmark between model-based planning and imitation learning, since the comparison is limited in scope and involves different experimental assumptions. Instead, the study highlights a complementary strength of online planning: the ability to adapt its motion generation to previously unseen geometric constraints by reasoning directly over the current scene. As a secondary observation, the same optimization pipeline used for dual UR5e experiments can be applied to a Franka-Panda platform without altering the underlying planning formulation.

\section{Conclusions and Future Work}\label{sec:conclusion}

We presented a sampling-based optimization framework for complex bimanual manipulation, leveraging a GPU-accelerated physics simulator as a high-fidelity world model. At the core of our approach is a customized sampling-based optimization that embeds a Quadratic Program within the sampling loop to enforce strict bounds on joint position, velocity, acceleration, and jerk. This projection decouples exploration from motion smoothness---two objectives that are fundamentally conflated in standard sampling-based optimizers---enabling the optimizer to search over the full kinematically reachable set while guaranteeing dynamically feasible trajectories. We derived a custom ADMM-based solver for this QP that exploits joint-wise separability and pre-computable matrix factorizations, making the projection negligibly cheap relative to the MuJoCo rollouts that dominate computation. We believe this decoupling principle extends beyond bimanual manipulation: any sampling-based MPC pipeline operating under kinematic or dynamic constraints can benefit from projecting samples onto the feasible set rather than restricting the sampling distribution to avoid violations.

We demonstrated the efficacy of our framework on four challenging bimanual tasks---tray moving, ball lifting and transport, cube handover, and box lifting---all involving contact-rich interactions in cluttered environments. Bayesian optimization of cost weights yielded substantial improvements in success rates across all tasks without additional computational overhead. Quantitative comparisons against baseline-CEM, low-pass filtering, and Savitzky--Golay smoothing confirmed that the QP-based projection achieves the most consistent performance, particularly in tasks requiring sustained bimanual coordination through obstacle-constrained workspaces. The framework operates at $10$--$16$\,Hz on a commodity laptop GPU and supports successful sim-to-real transfer on UR5e manipulators. We also presented a preliminary qualitative comparison with the learning-based approach of~\cite{gkanatsios20253d}, illustrating that online model-based planning and data-driven methods offer complementary strengths: our planner adapts to novel geometric configurations without retraining, while learning-based methods excel at mapping rich perceptual inputs to actions and amortizing computation across episodes.

Some limitations of the current framework are worth noting. First, while Bayesian optimization systematically tunes cost \emph{weights}, the \emph{structure} of the cost function---which terms to include, how to decompose the task into phases, and what geometric quantities to penalize---remains a manual design choice. Extending to significantly more complex tasks may require richer cost formulations or learned cost weights. Second, our framework relies on MuJoCo as a high-fidelity world model; tasks involving deformable objects, fluid interactions, or highly dynamic contacts may expose sim-to-real gaps that receding-horizon re-planning alone cannot compensate. 

These limitations point to several concrete directions for future work. The trajectories generated by our planner can serve as expert demonstrations for distilling a reactive policy network, which could in turn warm-start the optimizer in novel scenarios---reducing the number of samples and iterations required for convergence. Integrating a learned value function within a TD-MPC framework~\cite{hansen2022temporal} would allow our optimizer to reason over longer effective horizons without increasing the planning horizon $H$. The neural warm-starting strategy introduced in~\cite{andrejev2025pi} for the projection QP could further reduce computation time, potentially enabling higher control frequencies or larger batch sizes within the same time budget. Finally, extending the framework to tasks with dynamic obstacles, deformable objects, or tool use would test the generality of the modular cost design and the robustness of the sim-to-real transfer pipeline.

\section*{Acknowledgments}

The authors gratefully acknowledge the use of AI-assisted writing tools during the preparation of this manuscript. AI tools were used to assist with grammar correction, sentence restructuring, and clarity improvements in the writing. The AI system was used at the level of language refinement and copy-editing; all technical content, mathematical derivations, experimental design, results, and scientific conclusions were produced exclusively by the authors. No AI system was used to generate figures or data reported in this work.

\balance
\bibliography{references}
\bibliographystyle{IEEEtran}

% \newpage

 

% \end{IEEEbiography}


% \vfill

\end{document}